%% file: sample-sigconf-authordraft.tex
\PassOptionsToPackage{prologue,table}{xcolor}
\documentclass[sigconf, screen]{acmart}

\AtBeginDocument{%
  }

\setcopyright{acmlicensed}
\copyrightyear{2025}
\acmYear{2025}
\acmDOI{10.1145/XXXXXXX.XXXXXXX} 
\acmConference[ACMMM25 '25]{ACM Multimedia 2025, Dublin, Ireland}{October 27--31, 2025}{Dublin, Ireland}
\acmISBN{978-1-4503-XXXX-X/2025/10}
\acmSubmissionID{<YourSubmissionID>}

\usepackage{amsmath}
\usepackage{mathtools}
\usepackage{amsthm}
\usepackage[utf8]{inputenc} 
\usepackage[T1]{fontenc}    
\usepackage{hyperref} 
\usepackage{url}            
\usepackage{booktabs}       
\usepackage{amsfonts}       
\usepackage{nicefrac}       
\usepackage{microtype}      
\usepackage{xcolor}         
\usepackage{amssymb}
\usepackage{multirow}
\usepackage{enumitem}
\usepackage{fontawesome}
\usepackage{graphicx}
\usepackage{xspace}
\usepackage{graphicx}
\usepackage{amsmath}
\usepackage{amssymb}
\usepackage{booktabs}
\usepackage{algorithm}
\usepackage[table]{xcolor}
\usepackage{algorithmic}
\usepackage{multirow}
\newcommand{\ie}{{\emph{i.e.}}\xspace}
\usepackage[capitalize,noabbrev]{cleveref}
\theoremstyle{plain}
\newtheorem{theorem}{Theorem}[section]

\newtheorem{corollary}[theorem]{Corollary}
\theoremstyle{definition}
\newtheorem{definition}[theorem]{Definition}
\newtheorem{assumption}[theorem]{Assumption}
\theoremstyle{remark}

\acmSubmissionID{2523}

\begin{document}

\title{Free-Mask: A Novel Paradigm of Integration Between the Segmentation Diffusion Model and Image Editing}

\author{Bo Gao}
\affiliation{%
  \institution{Sun Yat-sen University}
  \city{Shenzhen}
  \country{China}}
\email{gaob7@mail2.sysu.edu.cn}

\author{Jianhui Wang}
\affiliation{%
  \institution{University of Electronic Science and Technology of China}
  \city{Chengdu}
  \country{China}}
\email{jianhuiwang@std.edu.uestc.cn}

\author{Xinyuan Song}
\affiliation{%
  \institution{Emory University}
  \city{Atlanta}
  \country{United States}
}
\email{xinyuan.song@emory.edu}

\author{Yangfan He}
\affiliation{%
  \institution{University of Minnesota-Twin Cities}
  \city{Minneapolis}
  \country{United States}
}
\email{he0005772@umn.edu}

\author{Fangxu Xing}
\affiliation{%
  \institution{Harvard Medical School}
  \city{Boston}
  \country{United States}
}
\email{fxing1@mgh.harvard.edu}

\author{Tianyu Shi}
\affiliation{%
  \institution{University of Toronto}
  \city{Toronto}
  \country{Canada}
}
\email{ty.shi@mail.utoronto.ca}
\renewcommand{\shortauthors}{Trovato et al.}
\settopmatter{printacmref=false} 
\begin{abstract}
Current semantic segmentation models typically require a substantial amount of manually annotated data, a process that is both time-consuming and resource-intensive. Alternatively, leveraging advanced text-to-image models such as Midjourney and Stable Diffusion has emerged as an efficient strategy, enabling the automatic generation of synthetic data in place of manual annotations. However, previous methods have been limited to generating single-instance images, as the generation of multiple instances with Stable Diffusion has proven unstable and masks can be significantly affected by occlusion between
different objects. To overcome this limitation and broaden the variety of synthetic datasets, we propose a novel framework, \textbf{Free-Mask}. It combines a Diffusion Model for segmentation with advanced image editing capabilities, allowing the insertion of multiple objects into images through text-to-image models. In addition, we introduce a new active learning paradigm that benefits both model generalization and data optimization. Our method enables the creation of realistic datasets that closely reflect open-world environments while generating accurate segmentation masks. Our code will be released soon.
\end{abstract}





\begin{teaserfigure}
  \includegraphics[width=\textwidth]{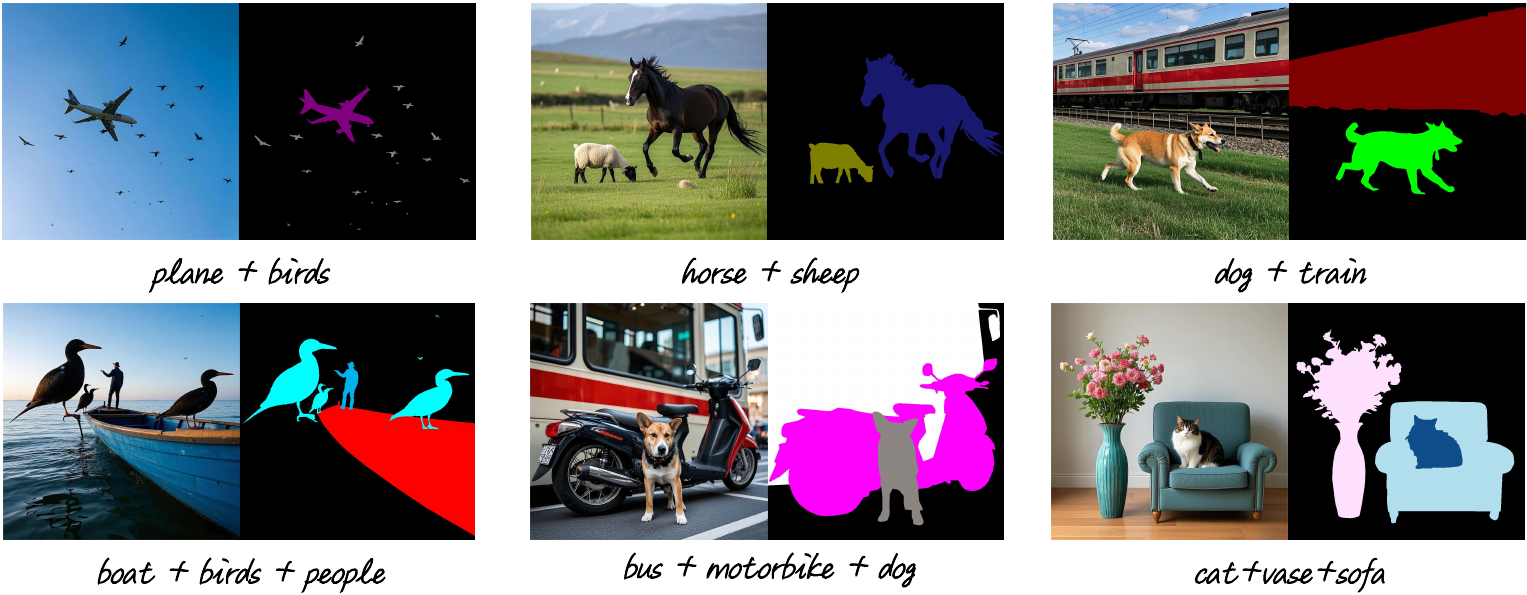}
  \vspace{-20pt}
  \caption{Free-Mask is capable of generating high-quality multi-instance segmentation results without the need for human annotations.
}
  \Description{Enjoying the baseball game from the third-base
  seats. Ichiro Suzuki preparing to bat.}
  \label{fig:teaser}
\end{teaserfigure}


\maketitle

\section{Introduction}
\label{intro}
Semantic segmentation~\cite{long2015fully,luo2019taking} is a critical computer vision component with profound implications across diverse applications. It provides detailed scene understanding, facilitating progress in autonomous driving, intelligent surveillance, and robotic navigation by interpreting complex visual environments. In addition, accurate segmentation strengthens object recognition and localization, improving tasks such as image retrieval and object identification. However, most semantic segmentation models still rely heavily on extensive, manually annotated data, which can be prohibitively expensive. For example, Cityscapes~\cite{cordts2016cityscapes} highlights that labeling a single semantic urban image can take about 60 minutes, underscoring the level of effort required.


To mitigate this issue, some previous works have explored weak supervision, which involves training models on rough annotations indicating only the presence or absence of certain object classes. Examples include image-level labels~\cite{kolesnikov2016seed,pei2022hierarchical,chen2021semantically,huang2018weakly,ahn2018learning,wei2018revisiting,jiang2019integral,chen2022saliency,li2022expansion,xu2023masked,huang2023paddles,huang2020low}, bounding boxes~\cite{khoreva2017simple,dai2015boxsup,jing2019coarse,lee2021bbam,song2019box}, points~\cite{bearman2016s}, and scribbles~\cite{lin2016scribblesup,vernaza2017learning}. These approaches aim to balance annotation costs with testing performance. Unfortunately, they still suffer from drawbacks such as low accuracy and complex training procedures, especially for tasks requiring precise masks.

\begin{figure}[t]
	\begin{center}
		\includegraphics[width=\linewidth]{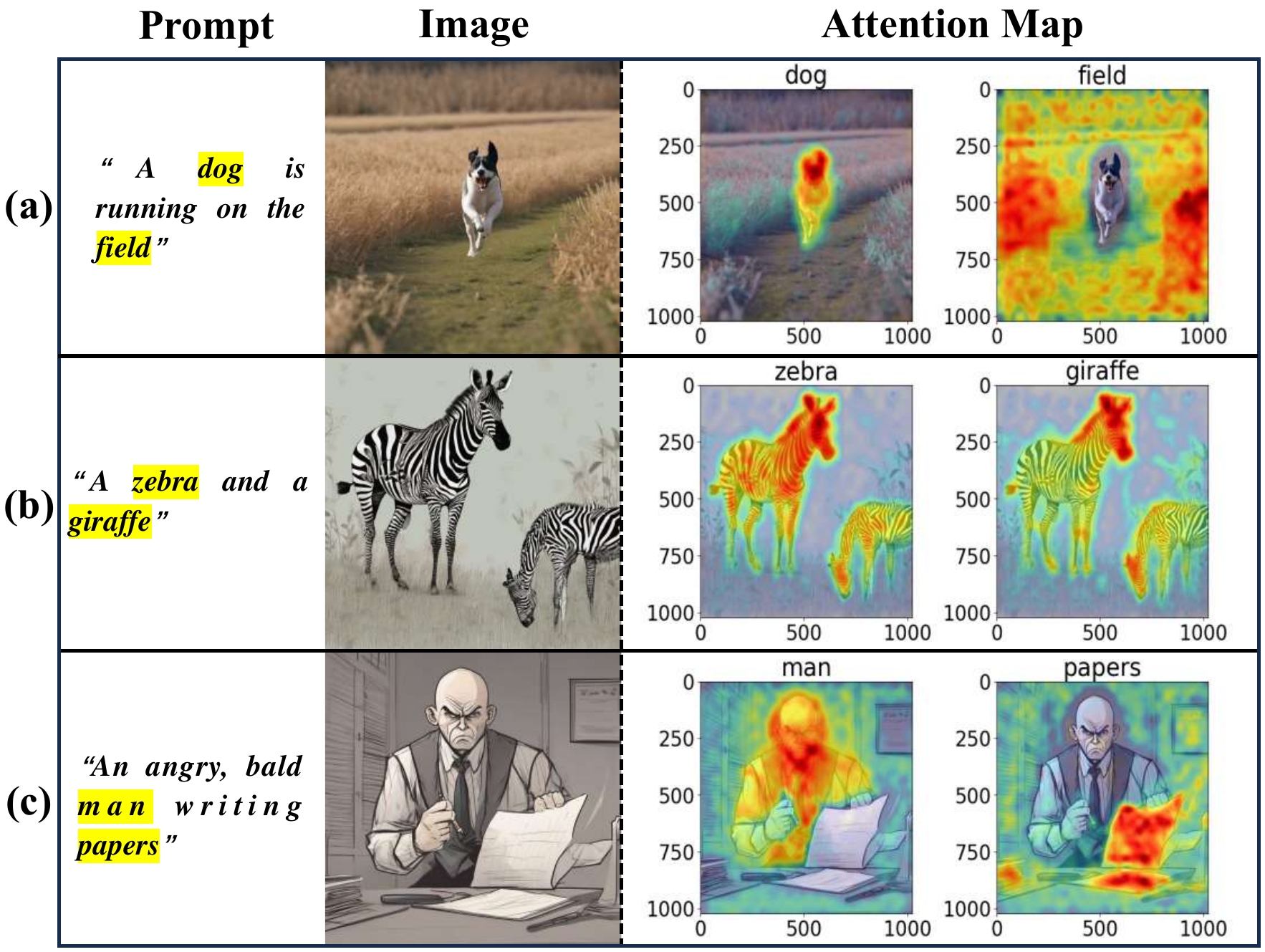}
	\end{center}
	\caption{\textcolor{red}{Why do conditioned diffusion models have the potential for generating masks?} (a) When the prompt is "A dog is running on the field," Stable Diffusion~\cite{rombach2022high} generates a high-quality image. Through DAAM~\cite{tang2023daam}, the attention map largely reflects correct semantic information for both the dog and the field. (b) Due to the limitations of Stable Diffusion~\cite{rombach2022high} in generating multi-instance images—especially for objects with similar semantics—it may lose some objects in the final image. (c) This is a successful multi-target generation. The attention maps titled “papers” and “man” respectively denote their specific regions and aims.}
	\label{attention map}
    \vspace{-10pt}
\Description{}\end{figure}

With the advancement of image generation technologies, diffusion models ~\cite{ramesh2022hierarchical,kawar2023imagic, Gao_2023_CVPR} have drawn considerable attention in the content generation community due to their stability and fidelity, surpassing GANs as displayed in ~\cite{Beat}. Especially for conditioned diffusion models, many cross attention layers exist within the architecture, and the output attention maps can be interpreted as coarse masks. As shown in Figure~\ref{attention map}, generating a heatmap (attention map) to visualize the image regions corresponding to each word in the text makes it clear that these maps can support more accurate results in later stages. Building on this technology, recent efforts such as DiffuMask~\cite{wu2023diffumask} and DiffusionSeg~\cite{ma2023diffusionseg} have produced synthetic images paired with segmentation masks. DiffuMask~\cite{wu2023diffumask} uses simple textual prompts such as “a photograph depicting a [class label]” to generate image-mask pairs. DiffusionSeg~\cite{ma2023diffusionseg} focuses on generating synthetic datasets aimed at object discovery by locating main objects in an image. However, these approaches are restricted to generating a single object segmentation mask per image. “Stable Diffusion is Unstable”~\cite{du2023stablediffusionunstable} indicates that the model still encounters certain issues (for example, \textit{Diffusion Constraint}), such as multi-target tasks shown in Figure~\ref{attention map} (b). Dataset Diffusion~\cite{quangtruong2023} can generate multiple objects along with accurate segmentation masks, though occlusion among different objects (\textit{Object-Occlusion Failure}) can reduce mask accuracy. Additionally, for DiffuMask~\cite{wu2023diffumask} and Dataset Diffusion~\cite{quangtruong2023}, specific thresholds must be selected (\ie{}, \textit{Empirical Threshold}). Typically, one can only choose the best threshold from a limited set or based on practical experience, making it challenging to achieve precise mask generation.


\begin{table*}[htbp]
	\centering
        \setlength\tabcolsep{9pt} 
	\caption{Comparisons between our approach and mainstream methods in terms of available ability and limitation. Only our method supports all generation modes and avoid some important flaws.}
        \resizebox{\linewidth}{!}
	{
            \renewcommand{\arraystretch}{1.5}
		\begin{tabular}{lccc}
			\hline
		  \textcolor{blue}{Ability}	& DiffuMask~\cite{wu2023diffumask}   & Dataset Diffusion ~\cite{quangtruong2023} & Free-Mask(ours) \\
			\hline
			 Single-target Generation & \faTimes & \faCheck & \faCheck  \\
	       Multiple-target Generation& \faTimes & \faCheck & \faCheck     \\
		   Free-Compose Generation & \faTimes & \faTimes & \faCheck   \\
                \hline
    
            \textcolor{red}{Limitation}  &  &  &  \\
            \hline
              Empirical Threshold & \faCheck & \faCheck & \faTimes   \\
              Diffusion Constraint & \faCheck & \faCheck & \faTimes   \\
              Object-Occlusion Failure &   & \faCheck  & \faTimes   \\
			\hline
		\end{tabular}
	}
	\label{tab:1_compare}
    \vspace{-10pt}
\end{table*}

\begin{figure*}[t]
	\begin{center}
		\includegraphics[width=0.9\textwidth]{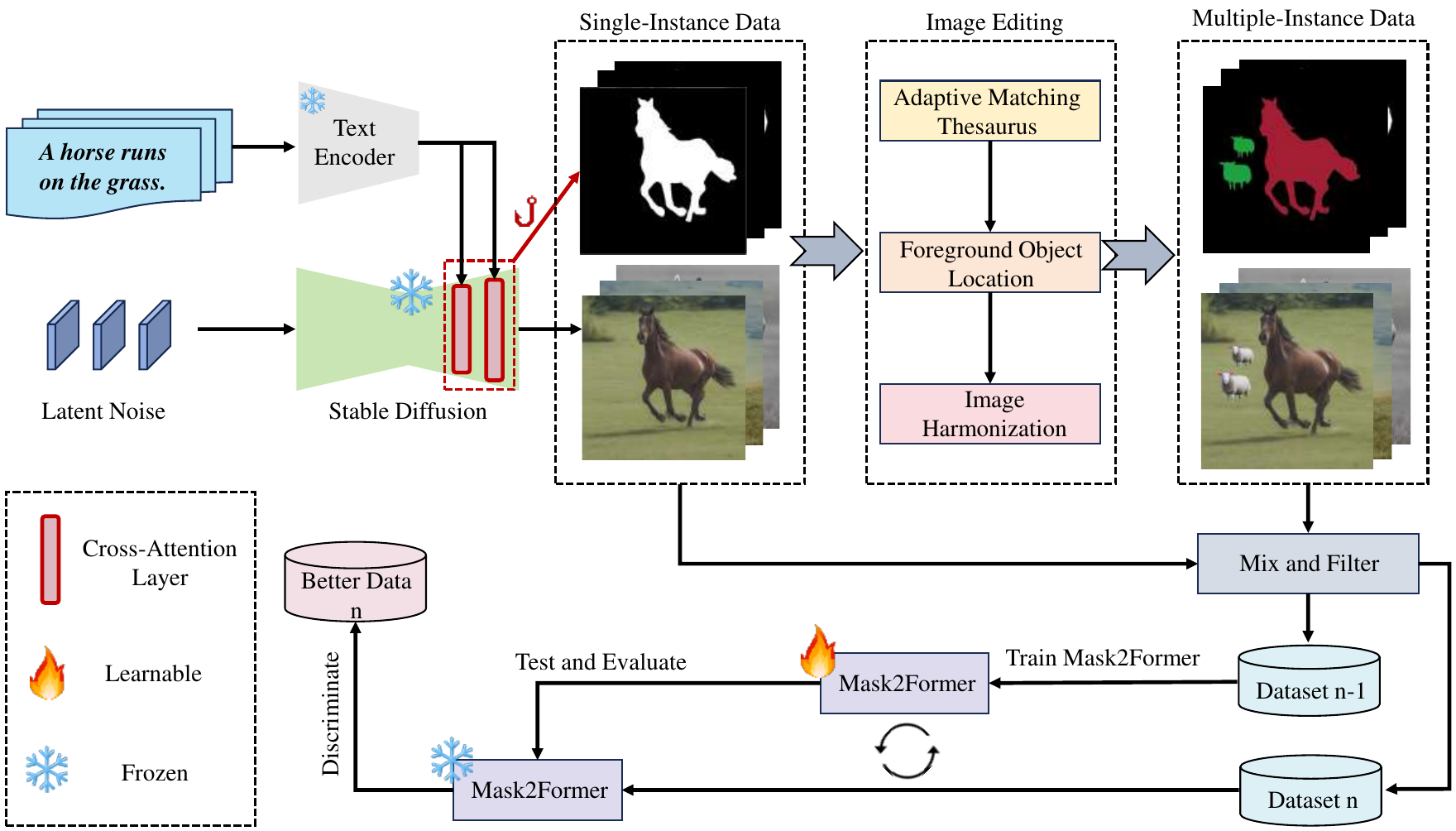}
	\end{center}
	\caption{Overall pipeline. Our goal is to generate semantic segmentation datasets for training and enhancing segmentation models. Unlike traditional approaches, we combine stable diffusion with image editing and layout diffusion models to create diverse datasets, including both single-object and multi-object scenarios.}
	\label{main task}  
\Description{}\end{figure*}
Based on the description provided (also outlined in Table~\ref{tab:1_compare}), we propose \textbf{Free-Mask}, an novel module for generating synthetic images and corresponding segmentation masks. Unlike previous methods such as DiffuMask~\cite{wu2023diffumask}, Free-Mask generates multiple objects along with their masks within a single image. Our approach uses the text-to-image diffusion model Stable Diffusion~\cite{rombach2022high}, which is trained on large web datasets. It can produce diverse images with fewer constraints from tightly related training data, providing a broad resource to improve the performance of segmentation models. In Free-Mask’s image editing pipeline, we employ three core techniques: 1) Adaptive Matching Thesaurus-selecting coherent and contextually appropriate objects to add to the scene; 2) Foreground Object Location-determining precise object placement locations within the image based on selections from the Adaptive Matching Thesaurus; and 3) Image Harmonization-integrating foreground objects seamlessly with the background to ensure visual consistency. These synthetic data are then utilized to train semantic segmentation methods, such as mask2former~\cite{cheng2022masked}, replacing real data and thus enhancing their robustness based on what we have proposed-Adversarial Active Iterative Learning, simultaneously enhancing the model's generalization ability and optimizing datasets in many application scenarios. Our main contributions are summarized as follows:
\begin{itemize}[left = 0em]

\item We present the \textbf{Free-Mask} framework, capable of generating multiple objects and single target in one image along with their accurate masks. This is the first work to combine a diffusion model for segmentation and image editing in a unified way. We have proposed a comprehensive generative mask pipeline that can handle diverse inputs and allows for various edits.

\item The data we generate includes both single-instance images and complex multi-instance images. We apply mixing and filtering to ensure that the synthesized data better matches real-world scenarios. We have conducted extensive experiments and achieved promising performance gains on two large-scale datasets and open-world scenes.

\item We introduce a new active learning approach that increases model generalization while refining the quality of selected data. This leads to higher-quality final datasets, improved model performance, and more efficient learning in practical applications.

\item We provide the theoretical analysis and key insights underlying our method. These results offer a rigorous foundation that supports the effectiveness and applicability of our framework design.
\end{itemize}

\setlength{\parskip}{0.1pt} 

\section{Related Work}
\subsection{Text-to-image Diffusion Models} 
Text-to-image diffusion models have become a notable approach for image generation. Since the publication of “Diffusion Models Beat GANs on Image Synthesis”~\cite{dhariwal2021diffusion}, many diffusion model variations have emerged for tasks such as image editing and super-resolution. For example, Imagic~\cite{kawar2023imagic} aligns a text embedding with both an input image and target text, then fine-tunes the model to capture image-specific features for image editing. IDM~\cite{Gao_2023_CVPR} proposes a denoising diffusion model with an implicit neural representation and a scale-controlled conditioning mechanism to address issues like over-smoothing and artifacts in continuous image super-resolution. According to DiffuMask~\cite{wu2023diffumask}, attention maps can be treated as mask annotations because the cross-attention layer is the sole pathway through which text influences image denoising, allowing the attention layers to focus on objects described in the text.

\subsection{Strongly and Weakly Supervised Semantic Segmentation} 
Due to the extensive manual annotation needed in strong supervision, many researchers have moved toward weakly supervised approaches. These techniques require only minimal annotations—points, bounding boxes~\cite{lee2021bbam}, or general classification tags~\cite{ahn2018learning,lee2021anti,wu2021embedded,xu2021leveraging,ru2022learning}—but they generally offer lower precision. Bounding boxes are more accurate but still demand considerable annotation, and most methods are limited to closed-set categories. To address these issues, approaches like DiffuMask~\cite{wu2023diffumask}, DiffusionSeg~\cite{ma2023diffusionseg}, ODISE~\cite{xu2023odise}, DiffSegmenter~\cite{wang2023diffusion}, and DifFSS~\cite{tan2023diffss} harness text-conditional diffusion models for weakly supervised segmentation. They rely on the correlation between text prompts and target objects through attention maps. These methods show strong potential by generating rich image-mask pairs.

\section{Methodology}
In generating datasets, the key shift in our approach (Figure~\ref{main task}) is from acquiring precise masks to editing images, facilitated by a precise and imposed locating scheme for mask annotation. In an open world with numerous objects, we face three primary challenges. First, it is essential to determine which objects can be appropriately added to generated images. For example, in an image of an airport produced by a diffusion model, it is logical to add airplanes but not giraffes. Second, it is crucial to decide where these objects should be placed within the image to ensure they fit the scene appropriately. For instance, a tree should be rooted on the ground rather than appearing to float in the air. Third, we must address discrepancies in physical conditions, such as lighting differences between the foreground objects and the background, to enhance the overall harmony.

To address these challenges, we propose a two-step strategy. Initially, we produce images with a single object per image and their corresponding masks.  Subsequently, we proceed to the second phase of image editing to tackle the aforementioned issues(Section~\ref{sec:3.2}).

\subsection{Single-object and Mask Generation}
\label{sec:3.1}
\subsubsection{Cross Attention Map in the Diffusion Model}
Text-to-Image models (e.g., Stable Diffusion~\cite{rombach2022high}, Imagen~\cite{saharia2022photorealistic}, DALLE·2~\cite{ramesh2022hierarchical}) are conditional diffusion models, where a text prompt $C_{text}$ guides the denoising process from the input Gaussian noise $z_t$ to the latent image $z_0$. Specifically, Stable Diffusion is composed of a text encoder $\varsigma_\theta$, a variational autoencoder (VAE), and a conditional UNet $\epsilon_\theta$. During generation, the text prompt $C_{text}$ is fed into cross-attention layers to be fused with visual latent embeddings. This fusion method allows the cross-attention map to highlight objects referenced in the text. 

For a given step $t$, assuming $z_t \in \mathbb{R}^{H \times W \times C}$, where $H$ is the height, $W$ is the width, and $C$ reflects the number of channels. It is linearly projected into the Query matrix $Q=ZW_q$. The text prompt $C_{text}$ is processed by the text encoder $\varsigma_\theta$, producing the text embeddings $X = \varsigma_\theta(C_{text}) \in \mathbb{R}^{L \times d}$. These embeddings are then linearly transformed into a Key matrix $K = XW_k$ and a Value matrix $V = XW_v$. Here, $W_q$, $W_k$, and $W_v$ are linear transformations for obtaining the query, key, and value. The cross-attention map at step $t$ and the $n_{th}$ layer of UNet is calculated as:
\begin{equation}
\mathcal{M}_C^{n,t}=\text{Softmax}\left(\frac{QK^T}{\sqrt{d}}\right),
\label{equ1}
\end{equation}
where $\mathcal{M}_C^{n,t} \in \mathbb{R}^{H \times W \times L}$ after projection, and $d$ is the dimensionality of the features.

\subsubsection{Mask Generation and Refinement}
Based on observations from~\cite{quangtruong2023}, using various timestep ranges only slightly affects the final outcome. Consequently, averaging these cross-attention maps across layers and timesteps is a straightforward approach for feature fusion, applied at four resolutions ($8\times8$, $16\times16$, $32\times32$, and $64\times64$), assuming an input size of $(512,512)$:
\begin{equation}
    \mathcal{M}_{C} = \frac{1}{N \times T} \sum_{n=1}^{N} \sum_{t=0}^{T} \mathcal{M}^{N,t}_{C}, \label{eq:average_attention}
\end{equation}
Although the cross-attention maps $\mathcal{M_C}$ provide an indication of the location of the target classes in the image, they often remain imprecise.
Following the approach of DiffuMask~\cite{wu2023diffumask}, which uses different classes of objects with different segmentation thresholds, we refine the vague masks. By utilizing AffinityNet ~\cite{ahn2018learning}, which helps identify object boundaries or perform pixel-level classification by learning the relationships or similarities between different regions in an image, we can generate multiple masks for foreground images of different classes based on varying thresholds. By comparing the intersection over union (IoU) with the output of  AffinityNet ~\cite{ahn2018learning}'s output, we can select the segmentation image that corresponds to the highest IoU value. This procedure provides an optimal segmentation mask for the relevant class.

\begin{figure}[t]
	\begin{center}
		\includegraphics[width=0.85\linewidth]{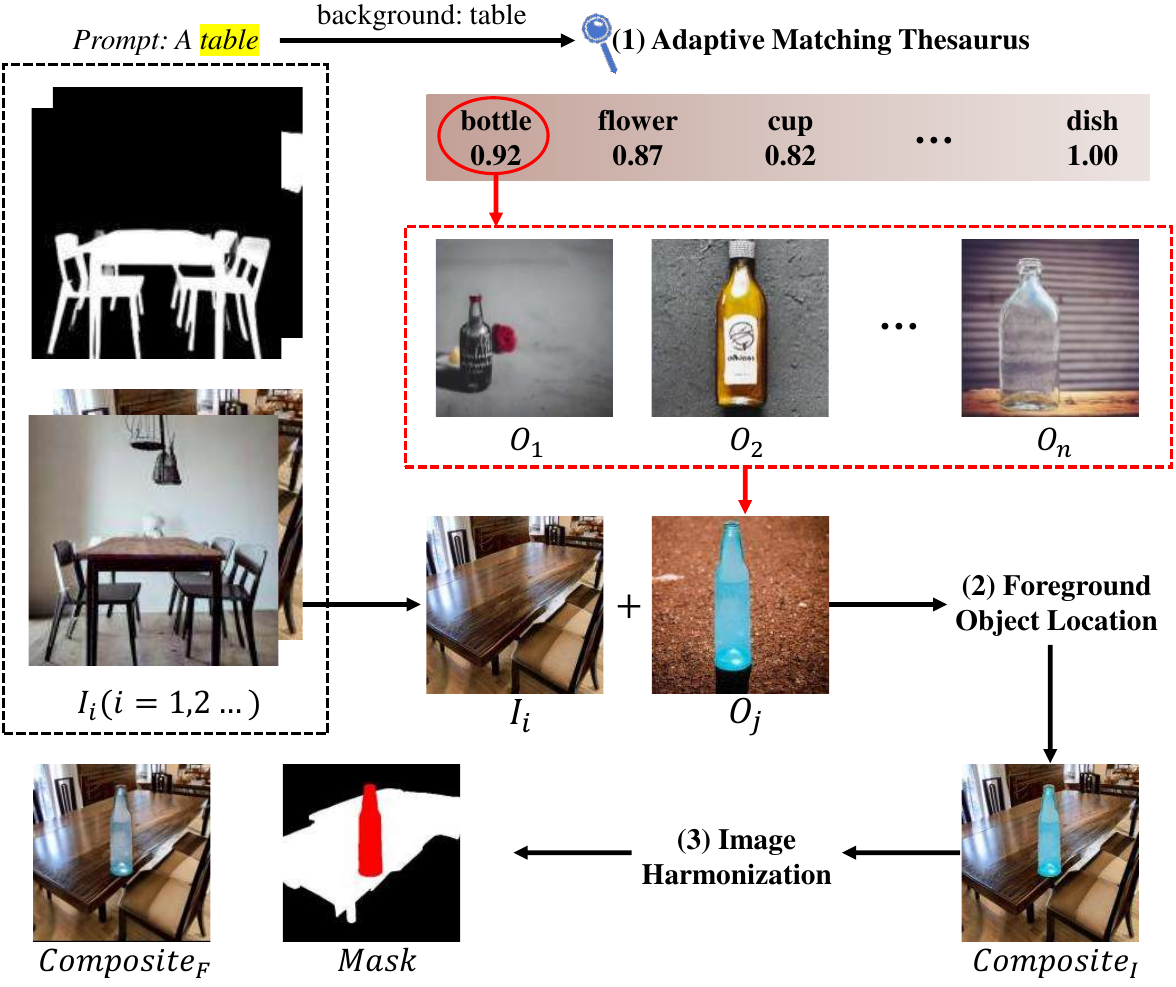}
	\end{center}
    \caption{IE pipeline: Overview of the image editing process.}
	\label{IE}
    \vspace{-10pt}
\Description{}\end{figure}

\subsection{Image Editing}
\label{sec:3.2}
As illustrated in Figure~\ref{IE}, our image editing pipeline consists of several key steps. First, we define the generated image set as \( I \) based on the single-instance dataset generated in the previous phase. We then construct the foreground target image set \( O \) using the adaptive matching thesaurus described below. For any image \( I_i \) and foreground target image \( O_j \), we obtain the location where the target should exist through object localization, thus generating the composite image \( Composite_I \). Since \( I_i \) and \( O_j \) come from images under different physical lighting conditions, we perform image harmonization modulation to ensure consistency. Finally, we produce a pair of data, \( Composite_F \) and \( Mask \).


\subsubsection{Adaptive Matching Thesaurus}
\label{sec: 3.2.1}
Stable Diffusion~\cite{rombach2022high} was trained on the large LAION dataset~\cite{schuhmann2022laion5b}, which contains diverse text-image pairs. Directly measuring semantic alignment between foreground and background objects from image features alone can be complicated and may risk overfitting. Therefore, an approach that starts with text-based analysis is both more straightforward and more efficient.

\begin{figure}[t]
    \centering
    \includegraphics[width=\linewidth]{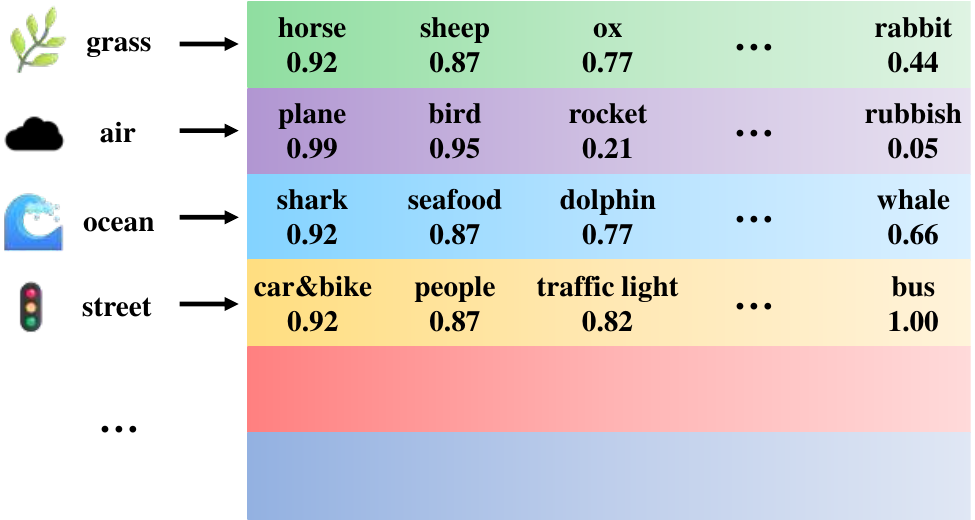}
    \caption{An example of the thesaurus built from text-image pairs.}
\label{fig: data}
\vspace{-10pt}
\Description{}\end{figure}

In this paper, we construct a broad matching lexicon to establish correspondences in an open-world environment by measuring the semantic similarity between foreground objects and backgrounds from a language understanding perspective. To quantitatively assess this matching, we calculate the frequency of foreground object terms appearing alongside given background references within a large text-image pair database. These probabilities serve as a measure of similarity. Specifically, the semantic relation \(\Tilde{\mathbb{R}}\) between an object \(o_i\) and a background \(b_j\) is computed as:

\begin{equation}
\Tilde{\mathbb{R}}(o_i,b_j) = \frac{\sum_{t \in \text{Text}_{b_j}} \mathbb{I}(o_i \in t)}{|\text{Text}_{b_j}|},
\end{equation}
where \(\text{Text}_{b_j}\) is the collection of prompts that include background \(b_j\), \(t\) is a single text prompt in \(\text{Text}_{b_j}\), and \(|\text{Text}_{b_j}|\) is the total number of those prompts. The indicator function \(\mathbb{I}(o_i \in t)\) returns 1 if the foreground object \(o_i\) appears in the prompt \(t\), otherwise 0. By using this formula, we build a semantic summary table for possible foreground-background pairs, as shown in Figure~\ref{fig: data}, making it easier to identify object categories suitable for embedding.

\subsubsection{Foreground Object Location} 
We follow the Fast Object Placement Assessment (FOPA)~\cite{niu2022fast} method to predict the optimal placement of a foreground object within a background. Let \( I_f \) be the foreground image, \( M_f \) be the binary segmentation mask of the foreground, and \( I_b \) be the background image. We employ a U-Net-like architecture consisting of an encoder-decoder structure. The encoder extracts feature representations \( F_{fg} \) and \( F_{bg} \) from the foreground and background images, respectively. The decoder produces a pixel-wise output feature map \( F_{out} \), where each pixel \( p_{ij} \) corresponds to the predicted score for placing the foreground object at position \( (i,j) \) within the background. The output at each pixel \( p_{ij} \), \( F_{out}(p_{ij}) \in [0, 1] \), indicates the confidence of the object being placed at that location.

\noindent\textbf{Complex Loss Function Design:} To further optimize the placement of the object, we introduce a more complex loss function that considers multiple aspects, including location accuracy, size matching, feature consistency, and heatmap alignment. (More detailed mathematics and the proof are in the supplementary material.)

\emph{Spatial Loss:} The spatial loss measures the combined error in the predicted location and size of the foreground object. We define the spatial parameters of the object as a 4-tuple \( \mathbf{P} = (c_x, c_y, w, h) \), where \( c_x \) and \( c_y \) denote the predicted center of the object, and \( w \) and \( h \) denote the predicted width and height. The ground truth parameters are denoted as \( \mathbf{P}_{\text{true}} = (c_{x, \text{true}}, c_{y, \text{true}}, w_{\text{true}}, h_{\text{true}}) \). A good bounding box regression loss function should consider three important geometric factors: overlap area, center point distance, and aspect ratio. Moreover, considering the hyperparameters, we normalize all the losses to a proportional loss, so that they all lie within the range $[0, 1]$.

\begin{equation}
\begin{split}
\mathcal{L}_{\text{spatial}} = &  \lambda_{\text{IoU}} (1 - \text{IoU}) \\
& + \lambda_{\text{center}} \left( 1 - \left( \frac{|c_x - c_{x, \text{true}}|}{c_{x, \text{true}}} + \frac{|c_y - c_{y, \text{true}}|}{c_{y, \text{true}}} \right) \right) \\
& + \lambda_{\text{aspect-ratio}} \left| \arctan\left( \frac{w}{h} \right) - \arctan\left( \frac{w_{\text{true}}}{h_{\text{true}}} \right) \right|,
\end{split}
\end{equation}

where \( \lambda_{\text{IoU}} \), \( \lambda_{\text{center}} \), and \( \lambda_{\text{aspect—ratio}} \) control the relative importance of the IoU loss, center point loss, and aspect ratio loss, respectively.

\emph{Semantic Loss:} The semantic loss measures how well the foreground object’s features align with the background’s semantic features. This is achieved by comparing the model’s output feature map \( F_{out}(p_i) \) with the feature representation \( F_{com}(i) \) of the composite image. By ensuring feature consistency, we improve the semantic integration of the foreground object into the background. The semantic loss is defined as:

\begin{equation}
\mathcal{L}_{\text{semantic}} = \sum_{(i,j)} \| F_{out}(p_{ij}) - F_{com}(p_{ij}) \|^2.
\end{equation}

\emph{Heatmap Loss:} The heatmap loss encourages the alignment of the predicted object placement with the true object placement. We define the discrepancy between the predicted heatmap \( H_{\text{pred}} \) and the ground truth heatmap \( H_{\text{true}} \) as:

\begin{equation}
\mathcal{L}_{\text{heatmap}} = \| H_{\text{pred}} - H_{\text{true}} \|_2^2.
\end{equation}

\emph{Unified Total Loss:}  The final total loss combines the spatial loss, semantic loss, and heatmap loss into a unified objective function. This ensures that the model optimizes the overall placement by balancing spatial accuracy, semantic consistency, and heatmap alignment. The total loss is expressed as:

\begin{equation}
\mathcal{L}_{\text{final}} = \lambda_1 \mathcal{L}_{\text{spatial}} + \lambda_2 \mathcal{L}_{\text{semantic}} + \lambda_3 \mathcal{L}_{\text{heatmap}}.
\end{equation}



\subsection{Adversarial Iterative Active Learning}
Inspired by active learning~\cite{Beluch_2018_CVPR,zhang2022,Kim_2023_ICCV,Mi_2022_CVPR}, we aim to train a segmentation model capable not only of learning from data but also of filtering and selecting data autonomously. We propose an adversarial iterative active learning framework that combines free-mask generation models with active learning strategies to optimize data quality and model performance. Specifically, we leverage our Free-Mask generation framework to produce an initial dataset containing samples of varying quality. Compared to traditional methods, we introduce a novel method: first, the generated data is used to train the Mask2Former model, which is then tested in real-world scenarios to evaluate its performance. Subsequently, the trained Mask2Former model serves as a discriminator to assess the quality of newly generated data from the framework. The bottom 30\% of samples with poor segmentation performance are discarded, while the top 70\% of high-quality samples are retained and added to the final dataset (Algorithm ~\ref{alg:active_learning}).
\begin{algorithm}[htbp]
\caption{Adversarial Active Iterative Learning Framework}
\label{alg:active_learning}
\begin{algorithmic}[1]
\REQUIRE Generator $G$, Initial model $f_{\theta_0}$, Training iterations $T$, Selection threshold $\alpha$
\STATE Initialize dataset $D_0 = G(z), \quad z \sim P_z$
\FOR{$t = 0$ to $T$}
    \STATE Train segmentation model $f_{\theta_t}$ on $D_t$:
    \STATE $\theta^*_t = \arg\min_{\theta} \sum_{(x, y) \in D_t} L(f_\theta(x), y)$
    \STATE Generate new data $D'_t = G(z), \quad z \sim P_z$
    \STATE Compute segmentation quality scores:
    \FOR{each $x \in D'_t$}
        \STATE $S(x) = \text{IoU}(f_{\theta^*_t}(x), y_{\text{pseudo}})$
    \ENDFOR
    \STATE Retain top $\alpha$-percentile of $D'_t$ based on $S(x)$:
    \STATE $D_{t+1} = \{ x \in D'_t \mid S(x) \geq S_{\alpha} \}$
\ENDFOR
\STATE \textbf{Return} Final dataset $D^*$ and trained model $f_{\theta^*}$
\end{algorithmic}
\end{algorithm}
\begin{table*}[!t]
    \caption{\textbf{Result of Semantic Segmentation on the VOC 2012 val.} And \textit{mIoU} is for $20$ classes. `S' and `R' refer to `Synthetic' and `Real'.}
    \centering
    \Large
    \setlength{\tabcolsep}{2pt}
    \resizebox{\linewidth}{!}{
    \input{tables/VOC_semantic.tex}
    }
    \label{VOC_semantic}

\end{table*}

This process iterates continuously, enhancing the dataset quality and optimizing Mask2Former’s segmentation capabilities through the interaction between the generation and discriminator models. As a result, we obtain a high-quality dataset while simultaneously training a high-performance segmentation model. Our experiments confirm the effectiveness of this framework in real-world settings, with significant improvements in data quality and model performance.

\section{Experiments and Analysis}
\subsection{Datasets and Metrics.}
We conduct evaluations on PASCAL VOC 2012~\cite{(voc)everingham2010pascal} and Cityscapes~\cite{(cityscape)cordts2016cityscapes}, two widely used semantic segmentation benchmarks. Performance is measured using mean Intersection-over-Union (\textbf{mIoU})~\cite{(voc)everingham2010pascal,cheng2022masked}. For open-vocabulary segmentation, following prior work~\cite{ding2022decoupling,cheng2021sign}, we report mIoU for seen and unseen classes, as well as their harmonic mean.

\subsection{Implementation Details}

\noindent\textbf{{Synthetic data for training.} }
Specifically, for Pascal-VOC 2012~\cite{(voc)everingham2010pascal}, we generate 10k images per category and filter out 7k images. Consequently, we assemble a final training set of 60k synthetic images across 20 classes, all with a spatial resolution of $512\times 512$ pixels. Regarding Cityscapes~\cite{cordts2016cityscapes}, we focus on evaluating two significant classes, namely 'Human' and 'Vehicle,' encompassing six sub-classes: \texttt{person}, \texttt{rider}, \texttt{car}, \texttt{bus}, \texttt{truck}, and \texttt{train}. We generate $30k$ images for each sub-category, with a final selection of 10k images per class.

\noindent\textbf{{The basic tools.} }
We leverage foundational components such as pre-trained Stable Diffusion~\cite{rombach2022high}, the text encoder from CLIP~\cite{radford2021learning}, and image harmonization techniques~\cite{ling2021region} as foundational components. And we utilize Mask2Former~\cite{cheng2022masked} as the baseline model for dataset evaluation. Without finetuning Stable Diffusion or training any module for individual categories, we maintain parameter optimization and settings consistent with the original papers, including initialization, data augmentation, batch size, and learning rate. All experiments are conducted using 8 Tesla A100 GPUs.

\subsection{Comparison with State-of-the-art Methods}

\subsubsection{Semantic Segmentation}
For the \textbf{VOC2012 dataset} ~\cite{(voc)everingham2010pascal}, as shown in Table~\ref{VOC_semantic}, our model outperforms the competition in nearly all of the 20 classes. Compared to DiffuMask ~\cite{wu2023diffumask}, when trained with purely synthetic data, our Free-Mask enhances the mIoU by nearly 8\% (from 57.4\% to 62.5\%) using Resnet50 ~\cite{he2016deep} and from 70.6\% to 72.0\% using Swin-B ~\cite{he2016deep}. Furthermore, our method also outperforms Dataset Diffusion~\cite{quangtruong2023}, achieving an mIoU improvement of 2\% (from 60.5\% to 62.5\%) using Resnet50 ~\cite{he2016deep}. In the "Finetune on Real Data" segment, which combines 60,000 synthetic images with 5,000 real images, the mIoU also increases by 9\%. Most notably, for categories such as "bird," "boat," "cat," "chair," and "sofa," our model demonstrates exceptionally strong performance, exceeding that achieved through training on real data by a significant margin, with an average gap exceeding 2\%.

Regarding the \textbf{Cityscapes dataset} ~\cite{cordts2016cityscapes}, our model continues to show robust capabilities. Compared to DiffuMask ~\cite{wu2023diffumask}, we have significantly reduced the performance gap with training on actual data, decreasing the difference from 10\% to single digits differences. Moreover, relying solely on generated data, We further boost results from about 70\% to over 80\% in purely synthetic data scenarios, as illustrated in Table~\ref{cityscapes_human}.

\begin{table}[t]
    \caption{\textbf{The mIoU (\%) of Semantic Segmentation on Cityscapes val}. `Human' includes two sub-classes {person} and rider. `Vehicle' includes four sub-classes, ~\ie \ {car}, {bus}, {truck}, and {train}. Mask2former~\cite{cheng2022masked} with ResNet50 is used.}
    \centering
    \footnotesize
    \renewcommand{\arraystretch}{1.0}
    \resizebox{\linewidth}{!}{
        \input{tables/cityscapes_human_car.tex}
    }
    \label{cityscapes_human}
\end{table}

\begin{table}[t]
    \caption{\textbf{Performance for Zero-Shot Semantic Segmentation Task on PASCAL VOC.} `Seen,' `Unseen,' and `Harmonic' denote mIoU of seen, unseen categories, and their harmonic mean. Priors are trained with real data and masks.}
    \centering
    \normalsize
    \renewcommand{\arraystretch}{1.3}
    \resizebox{\linewidth}{!}{
        \input{tables/zs3.tex}
    }
    \label{zs3}
    \vspace{-10pt}
\end{table}

\subsubsection{Zero-Shot Segmentation}
Even without fine-tuning, our model maintains open-world segmentation capabilities. As indicated in Table~\ref{zs3}, compared to many previous methods trained exclusively on real images and manually annotated masks, it achieves state-of-the-art results in zero-shot scenarios. All data automatically generated by Free-Mask are synthetic, eliminating the laborious manual annotation and dataset construction processes. We have also expanded the scope from single objects to multiple instances, enriching the dataset without introducing significant additional effort. Even when using the COCO dataset ~\cite{lin2014microsoft}to predict pseudo labels, which incurs higher computational costs, our approach, relying only on synthetic data, achieves a promising result of 66.6\% on unseen classes, representing an almost 3\% improvement over previous methods. More visualization results are in Appendix.

\subsubsection{Visual Comparison}
\begin{figure*}[htbp]
    \centering
    \includegraphics[width=\linewidth]{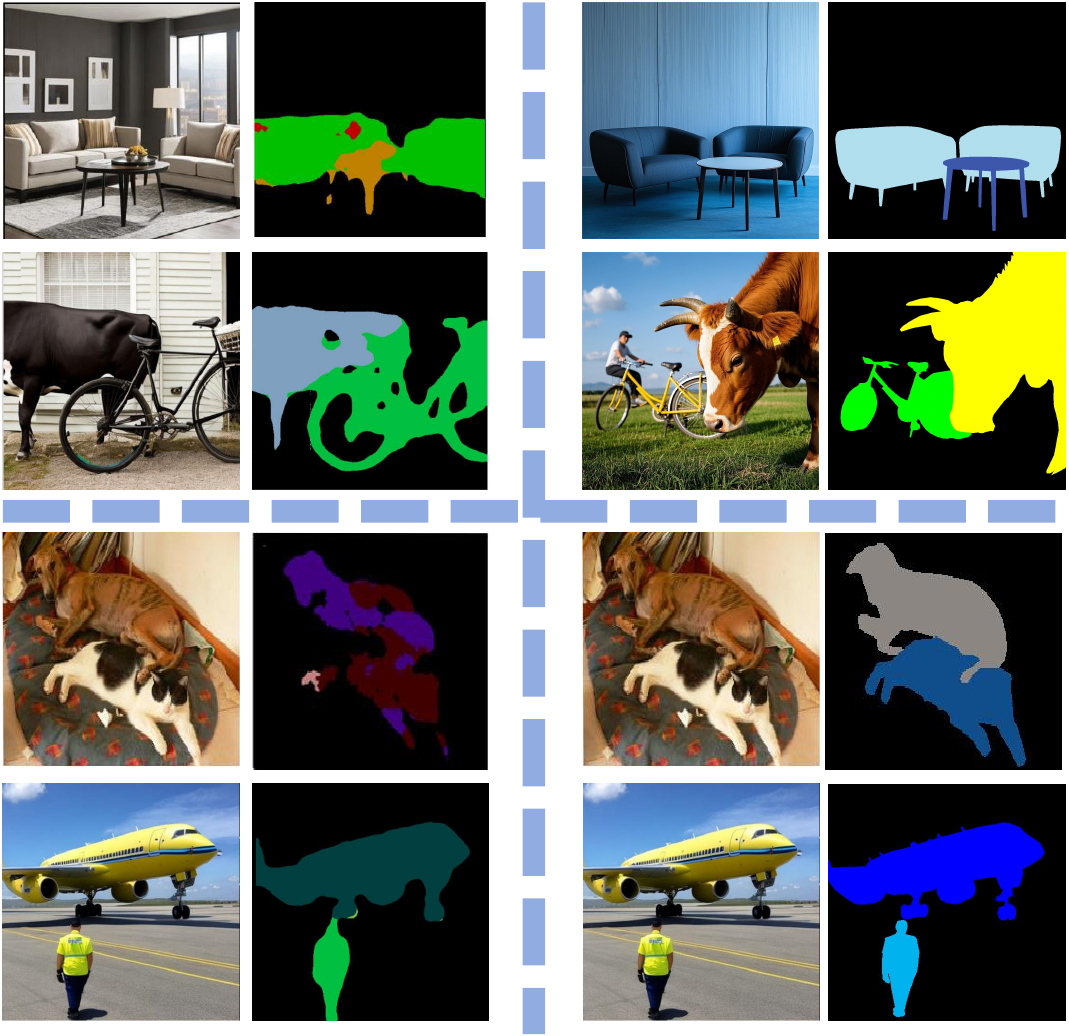}
    \caption{Comparison of mask accuracy under occlusion. The top-left shows masks using a generate-then-segment approach (Dataset Diffusion~\cite{quangtruong2023} or SAM~\cite{kirillov2023segany}), which can lose accuracy when objects overlap. Our method (bottom row) adds objects directly to the original scene and updates masks on occluded regions consistently, leading to significant improvements in handling occlusions.}
\label{fig:compare}
\vspace{-10pt}
\Description{}\end{figure*}

Figure~\ref{fig:compare} displays results from two totally different ideas. The top-left corner shows results from Dataset Diffusion~\cite{quangtruong2023} and the masks obtained by applying SAM~\cite{kirillov2023segany} on our diffusion-generated images. This approach is, on one hand, limited by the diffusion model's capability to handle multi-instance generation, as demonstrated in the ~\ref{intro}. We also observe that in complex scenes, object occlusion leads to highly inaccurate masks when using this generate-then-segment approach. In contrast, our method effectively avoids the inaccuracies caused by occlusion when adding objects to an image. This is because we add new objects on top of the original foreground and background. If occlusion occurs, the newly added object naturally overlaps part of the original object, and the new mask directly overwrites the original one, ensuring consistency between the two. As a result, the outcomes (COCO 2017~\cite{caesar2018cocostuffthingstuffclasses}) in the bottom-left and bottom-right corners also demonstrate that our method handles occlusions significantly better.

\subsection{Ablation Study}

\begin{figure*}[htbp]
    \centering
    \includegraphics[width=\linewidth]{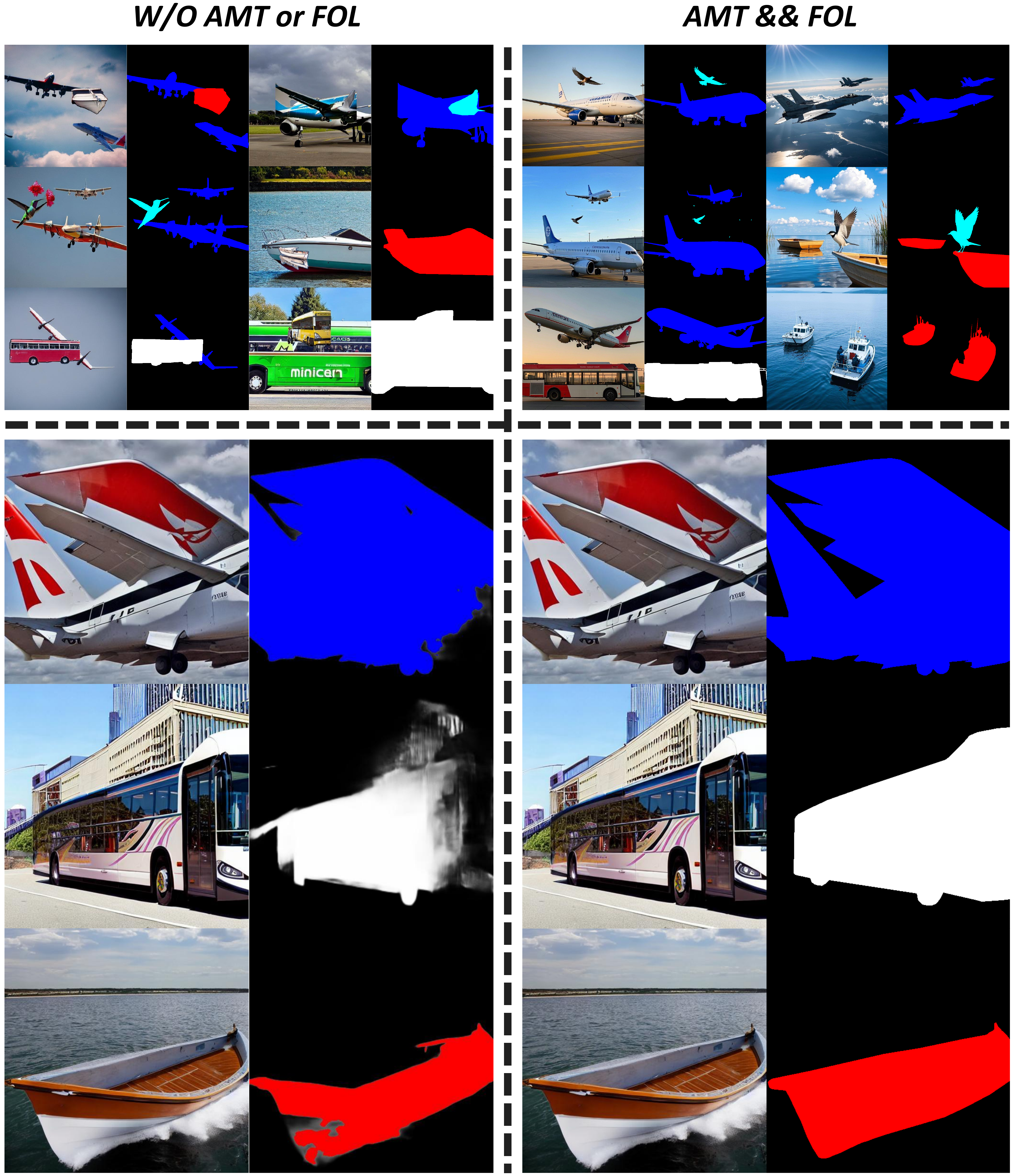}
    \caption{The top-left quadrant shows raw data without any of the two refinement methods. The top-right quadrant presents data refined with our method. The bottom-left quadrant displays results obtained by training on raw data, while the bottom-right quadrant shows the outcome after training with our method. This comparison demonstrates that the Adaptive Matching Thesaurus (AMT) and Foreground Object Location (FOL) noticeably improve data accuracy. For instance, when two buses of the same category appear, a model lacking precise localization can mistakenly interpret a protruding part of one bus as belonging to the other, leading to severe segmentation errors. Moreover, without AMT, visually similar but semantically unrelated objects can confuse boundaries and produce poor segmentation performance, as shown in the bottom-left quadrant.
    }
\label{fig: ablation}
\Description{}\end{figure*}

\subsubsection{Role of Adaptive Matching Thesaurus} 

As discussed in Section~\ref{sec: 3.2.1}, the thesaurus plays a vital role in filtering unfit objects semantically. In Table~\ref{gap}, a 2.6\% gap can be observed for all classes on average. This gap confirms the importance of semantic consistency in semantic segmentation.

\begin{table}[htbp]
   \caption{Ablation Study: The results are presented on the VOC 2012 validation dataset, achieved when fine-tuning on real-world data, with the average mIoU(\%) for all classes.}
    \centering
    \setlength{\tabcolsep}{1pt}
    \input{tables/Ablation_GaP}
    \label{gap}
\end{table}

\subsubsection{Effectiveness of Foreground Object Location} 

Without Foreground Object Location, we could not get nearly realistic images. Even in a simple photo of two airplanes in the sky, if there isn't an appropriate positioning, the airplanes might overlap. Such dirty data can significantly interfere with the training of segmentation models and affect their performance. Thus, there is a 1.8\% gap.

\subsubsection{Visualization of Data Refinement}
Figure~\ref{fig: ablation} provides an overview of the improvements gained by our data refinement techniques. The refined data lead to noticeably better segmentation performance. More details and illustrations on each quadrant are provided in the figure caption.

\subsubsection{Effectiveness of Interactive Active Learning} 
\begin{figure}[htbp]
	\begin{center}
		\includegraphics[width=0.9\linewidth]{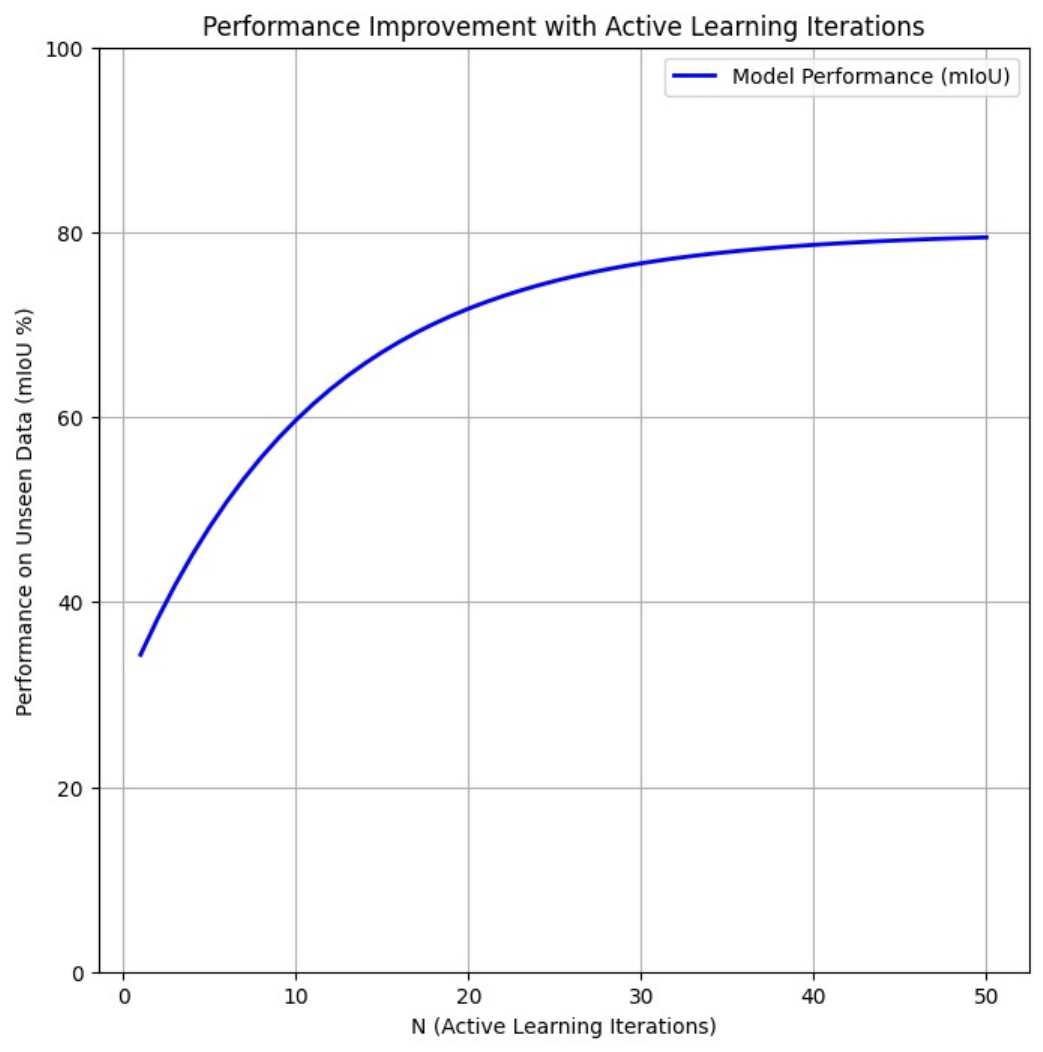}
	\end{center}
	\caption{Effectiveness of Interactive Active Learning: We use mIoU to assess performance with different iterations (N) on the VOC 2012 validation dataset.}
	\label{Interactive Activate Learning}
-    \vspace{-15pt}
\end{figure}

By combining free-mask generation with active learning strategies, our method raises segmentation accuracy from 68\% to nearly 79\%. The framework automatically filters low-quality samples (the bottom 30\%) and retains the top 70\%, ensuring high-quality training data. Through iterative refinement, both dataset quality and model performance steadily improve, making the approach robust for handling noisy, real-world data.

In Figure~\ref{Interactive Activate Learning}, the model initially benefits significantly from newly collected high-information data, which speeds up learning and boosts the mIoU. However, as iterations progress, the model becomes more refined,  and the incremental gains from additional data become smaller. This diminishing return effect is a common characteristic of active learning, where the most informative samples are selected early, and later iterations require substantially more data to achieve smaller performance gains.

\section{Conclusion}

Building upon our framework \textbf{Free-Mask}, this paper removes the reliance on manual annotation for semantic segmentation by integrating image editing with generative models. Unlike previous methods that only produce single-instance images, our approach can handle multiple instances within a single image, improving its applicability to various real-world scenarios. In addition, the new active learning strategy refines data quality and strengthens model generalization, enabling the creation of more realistic datasets that reflect open-world conditions while generating accurate segmentation masks. Moving forward, we plan to further extend this method by deriving multi-instance, multi-object segmentation masks directly from the Diffusion Model, achieving an end-to-end augmentation pipeline. Our code will be released soon.

\bibliographystyle{ACM-Reference-Format}
\bibliography{sample-sigconf-authordraft}

\clearpage
\appendix

\section{Theorem 1: Approximation Property of Cross-Attention Maps to Semantic Masks}

First, we want to make the basic assumptions and definitions for our theory.

\subsection{Definitions and Assumptions}
\begin{definition}[Diffusion Model]
Let the UNet in Stable Diffusion have \(N\) cross-attention layers. The attention map at layer \(n\) and timestep \(t\) is denoted as \(\mathcal{M}_C^{n,t} \in \mathbb{R}^{H \times W \times L}\), where \(H \times W\) is the spatial resolution and \(L\) is the length of the text prompt.
\end{definition}

\begin{assumption}[Text-Image Alignment]
Given a text prompt \(C_{\text{text}} = \{w_1, \dots, w_L\}\), the true semantic mask \(\text{Mask}_{\text{true}} \in \{0,1\}^{H \times W}\) satisfies:
\begin{equation}
\text{Mask}_{\text{true}}(i,j) = 1 
\quad\Longleftrightarrow\quad 
\text{pixel } (i,j) \text{ belongs to object } w_k.
\end{equation}
\end{assumption}

\begin{definition}[Attention Map Aggregation]
The averaged attention map \(\mathcal{M}_C\) is defined as:
\begin{equation}
\mathcal{M}_C 
= \frac{1}{N \times T} \sum_{n=1}^N \sum_{t=0}^T \mathcal{M}_C^{n,t},
\end{equation}
where \(\mathcal{M}_C^{n,t}\) is the attention map at layer \(n\) and timestep \(t\), \(N\) is the total number of cross-attention layers, and \(T\) represents the number of timesteps.
\end{definition}

\begin{definition}[Binarization]
For a given threshold \(\tau\), the binarized mask \(\hat{\text{Mask}}\) is defined as:
\begin{equation}
\hat{\text{Mask}}(i,j) 
= \mathbb{I}\Bigl(\mathcal{M}_C(i,j) \ge \tau\Bigr),
\end{equation}
where \(\mathbb{I}(\cdot)\) is the indicator function.
\end{definition}

The we can get our main theorem:

\begin{theorem}[IoU Lower Bound of Binarized Attention Map]
\label{thm:IoU_bound}
Suppose the text-image alignment strength \(\alpha\) satisfies
\begin{equation}
\alpha := \min_{w_k \in C_{\text{text}}} P\Bigl(\mathcal{M}_C(i,j \mid w_k) \geq \tau \Bigm\vert \text{Mask}_{\text{true}}(i,j)=1\Bigr) > 0.5.
\end{equation}
Then the IoU between the binarized attention map \(\hat{\text{Mask}}\) and the true mask \(\text{Mask}_{\text{true}}\) satisfies the following lower bound:
\begin{equation}
\begin{aligned}
&\text{IoU}(\hat{\text{Mask}}, \text{Mask}_{\text{true}}) \geq \frac{2\alpha - 1}{\alpha + \beta}, \\
&\text{where} \quad 
\beta = \max_{w_k \in C_{\text{text}}} P\Bigl(\mathcal{M}_C(i,j \mid w_k) \geq \tau \Bigm\vert \text{Mask}_{\text{true}}(i,j)=0\Bigr).
\end{aligned}
\end{equation}
\end{theorem}

\subsection{Detailed Proof of Theorem~\ref{thm:IoU_bound}}

We start by applying the law of total probability to the event that the binarized mask is active at pixel $(i,j)$:

\begin{equation}
\begin{aligned}
\mathbb{P}\Big(\hat{\text{Mask}}(i,j)=1\Big) &= \mathbb{P}\Big(\hat{\text{Mask}}(i,j)=1\mid\text{Mask}_{\text{true}}(i,j)=1\Big) \\
&\cdot \mathbb{P}\Big(\text{Mask}_{\text{true}}(i,j)=1\Big)\\
&\quad + \mathbb{P}\Big(\hat{\text{Mask}}(i,j)=1\mid\text{Mask}_{\text{true}}(i,j)=0\Big) \\
&\cdot \mathbb{P}\Big(\text{Mask}_{\text{true}}(i,j)=0\Big).
\end{aligned}
\end{equation}

By definition, we have the following conditional probabilities:
\begin{equation}
\begin{aligned}
\mathbb{P}\Big(\hat{\text{Mask}}(i,j)=1\mid\text{Mask}_{\text{true}}(i,j)=1\Big) &= \alpha,\\begin{equation}1mm]
\mathbb{P}\Big(\hat{\text{Mask}}(i,j)=1\mid\text{Mask}_{\text{true}}(i,j)=0\Big) &= \beta.
\end{aligned}
\end{equation}

Let the true mask coverage be denoted by
\begin{equation}
\mathbb{P}\Big(\text{Mask}_{\text{true}}(i,j)=1\Big) = p,
\end{equation}
which implies that
\begin{equation}
\mathbb{P}\Big(\text{Mask}_{\text{true}}(i,j)=0\Big)=1-p.
\end{equation}

We get

\begin{equation}
\begin{aligned}
\mathbb{P}\Big(\hat{\text{Mask}}(i,j)=1\Big) &= \alpha \cdot p + \beta \cdot (1-p) .
\end{aligned}
\end{equation}

The IoU can be expressed as:
\begin{equation}
\begin{aligned}
\text{IoU} &= \frac{\mathbb{P}\Big(\hat{\text{Mask}}(i,j)=1 \land \text{Mask}_{\text{true}}(i,j)=1\Big)}{\mathbb{P}\Big(\hat{\text{Mask}}(i,j)=1 \lor \text{Mask}_{\text{true}}(i,j)=1\Big)} \\ &= \frac{\alpha p}{\alpha p + \beta (1-p) + p - \alpha p} \\&= \frac{\alpha p}{p + \beta (1-p)}.
\end{aligned}
\end{equation}

To find the worst-case lower bound, minimize IoU w.r.t. \(p\):
\begin{equation}
\frac{d}{dp} \Big( \frac{\alpha p}{p + \beta (1-p)} \Big) = \frac{\alpha \beta}{(p + \beta (1-p))^2} > 0.
\end{equation}
Thus, IoU attains its minimum as \(p \to 0\):
\begin{equation}
\lim_{p \to 0^+} \text{IoU} = \frac{2\alpha - 1}{\alpha + \beta}.
\end{equation}
Note that when \(\alpha > 0.5\) and \(\beta < \alpha\), the denominator \(\alpha + \beta < 2\alpha - 1\) ensures a positive lower bound.

We now provide a formal justification for why optimizing the threshold $\tau$ via AffinityNet leads to an improved IoU by (i) increasing the true‐positive rate $\alpha$ and (ii) decreasing the false‐positive rate $\beta$.

\begin{equation}
\text{IoU}(\tau) = \frac{\alpha(\tau)p}{p+\beta(\tau)(1-p)}.
\end{equation}

In our formulation the quantities $\alpha$ and $\beta$ depend on the choice of threshold $\tau$. In particular, define

\begin{equation}
\begin{aligned}
\alpha(\tau) &= P\Big(\mathcal{M}_C(i,j) \ge \tau \Big| \text{Mask}_{\text{true}}(i,j)=1\Big),\\begin{equation}1mm]
\beta(\tau) &= P\Big(\mathcal{M}_C(i,j) \ge \tau \Big| \text{Mask}_{\text{true}}(i,j)=0\Big).
\end{aligned}
\end{equation}

A lower threshold $\tau$ generally increases both $\alpha$ and $\beta$, while a higher threshold decreases them. The optimal $\tau$ should strike a balance so as to maximize IoU. In other words, we choose

\begin{equation}
\tau^* = \mathop{\rm argmax}_\tau \text{IoU}(\tau)=\mathop{\rm argmax}_\tau \frac{\alpha(\tau)p}{p+\beta(\tau)(1-p)}.
\end{equation}

AffinityNet is a network that learns the spatial affinities between pixels. When used to optimize $\tau$, it performs the following two roles: (1) It encourages better localization by promoting consistency among neighboring pixels. This affinity enforces that pixels truly belonging to the object have similar attention values. As a result, for the true object regions the conditional probability $\alpha$($\tau$) increases for an appropriately chosen $\tau$. (2) It suppresses spurious activations in the background by penalizing inconsistent regions. In turn, the probability $\beta$($\tau$) decreases when $\tau$ is optimized.

Suppose that the AffinityNet loss $L_{aff}$ is designed (and trained) so that at the optimal threshold $\tau^*$ we have

\begin{equation}
\frac{\partial \alpha(\tau)}{\partial \tau}\Big|_{\tau=\tau^*} > 0 \quad\text{and}\quad \frac{\partial \beta(\tau)}{\partial \tau}\Big|_{\tau=\tau^*} < 0,
\end{equation}

meaning that small changes in $\tau$ away from $\tau^*$ would degrade the performance. Then, at $\tau^*$ the IoU function

\begin{equation}
\text{IoU}(\tau)=\frac{\alpha(\tau)p}{p+\beta(\tau)(1-p)}
\end{equation}

is maximized. In other words, by differentiating IoU with respect to $\tau$ we identify

\begin{equation}
\frac{\partial \text{IoU}(\tau)}{\partial \tau}\Big|_{\tau=\tau^*} = 0,
\end{equation}

and the second-order condition (together with the network’s affinity constraints) guarantees that $\tau^*$ is a local maximum. The improved IoU bound is then given by

\begin{equation}
\text{IoU}(\tau^*)\ge \frac{2\alpha(\tau^*)-1}{\alpha(\tau^*)+\beta(\tau^*)},
\end{equation}

The effect of the AffinityNet optimization is that at $\tau^*$ we obtain a higher $\alpha$ and a lower $\beta$ than if $\tau$ were chosen arbitrarily. This directly “tightens” the lower IoU bound.

\begin{corollary}
If multi-scale attention maps (\(8\times 8\) to \(64\times 64\)) exhibit hierarchical consistency such that \(\alpha\) increases with resolution, then the following inequality holds:
\begin{equation}
\text{IoU} \geq \epsilon 
\quad \text{with} \quad 
\epsilon \to 1 \ \text{as} \ N, T \to \infty.
\end{equation}
\end{corollary}

Consider a theoretical result that assumes multi-scale attention maps (ranging from \(8\times 8\) to \(64\times 64\)) can capture increasingly detailed patterns as resolution grows. This assumption implies that the attention maps become more reliable at distinguishing object boundaries, thereby improving the overall segmentation accuracy. The corollary then provides a direct implication of this assumption by stating that if the attention maps exhibit hierarchical consistency across these scales, a parameter \(\alpha\) (which measures the reliability of the segmentation) improves with resolution. Consequently, the intersection-over-union (IoU) metric can be bounded below by a value \(\epsilon\) that approaches 1 as the number of cross-attention layers \(N\) and timesteps \(T\) approach infinity.  

In simpler terms, when the our model incorporates increasingly fine-grained attention information, any initial segmentation imprecision is systematically corrected in higher-resolution layers, leading to better alignment between predicted and true masks. The corollary formalizes this process by quantifying the rate at which the IoU approaches its optimal value. As a result, under ideal conditions (sufficiently large \(N\) and \(T\)), the predicted masks converge toward near-perfect segmentation, indicating a high level of agreement with the true masks.

\section{Theorem 2: Optimality of Foreground Object Placement
 Statement}
\begin{definition}[Background and Foreground Images]
Let the background image be \(I_b \in \mathbb{R}^{H \times W \times 3}\), and let the foreground object image be \(I_f \in \mathbb{R}^{h \times w \times 3}\) with its true mask \(M_f \in \{0,1\}^{h \times w}\). 
\end{definition}

\begin{definition}[Object Placement Parameters]
Define the object placement parameters as \(\mathbf{P} = (c_x, c_y, w, h)\), where \((c_x, c_y)\) are the center coordinates for placing \(I_f\) in \(I_b\), and \(w, h\) are the width and height of the foreground region.
\end{definition}

\begin{definition}[Composite Loss Function]
Let the final composite loss \(\mathcal{L}_{\text{final}}\) be given by:
\begin{equation}
\mathcal{L}_{\text{final}} 
= \lambda_1 \mathcal{L}_{\text{spatial}} 
+ \lambda_2 \mathcal{L}_{\text{semantic}} 
+ \lambda_3 \mathcal{L}_{\text{heatmap}},
\end{equation}
where \(\lambda_1 + \lambda_2 + \lambda_3 = 1\). The individual terms are defined as follows:
\begin{itemize}[left = 0em]
    \item Spatial Loss: \(\mathcal{L}_{\text{spatial}}\): 
    \begin{equation}
    \begin{aligned}
    \mathcal{L}_{\text{spatial}} 
    &= \lambda_{\text{IoU}}(1 - \text{IoU})
    + \lambda_{\text{center}} 
    \Bigl(1 
    - \frac{\lvert c_x - c_{x,\text{true}}\rvert}{c_{x,\text{true}}} 
    - \frac{\lvert c_y - c_{y,\text{true}}\rvert}{c_{y,\text{true}}}\Bigr) \\
    &\quad 
    + \lambda_{\text{aspect-ratio}} 
    \bigl\lvert
    \arctan\bigl(\tfrac{w}{h}\bigr)
    - \arctan\bigl(\tfrac{w_{\text{true}}}{h_{\text{true}}}\bigr)
    \bigr\rvert.
    \end{aligned}
    \end{equation}
    \item Semantic Loss: \(\mathcal{L}_{\text{semantic}}\):
    \begin{equation}
    \mathcal{L}_{\text{semantic}} 
    = \sum_{(i,j)} 
    \|F_{\text{out}}(p_{ij}) 
    - F_{\text{com}}(p_{ij})\|^2,
    \end{equation}
    where \(F_{\text{out}}\) is the predicted feature map, and \(F_{\text{com}}\) is the feature map of the composite image.
    \item Heatmap Loss: \(\mathcal{L}_{\text{heatmap}}\):
    \begin{equation}
    \mathcal{L}_{\text{heatmap}}
    = \|H_{\text{pred}} 
    - H_{\text{true}}\|_2^2.
    \end{equation}
\end{itemize}
\end{definition}

\begin{definition}[Physical Plausibility Score]
Define the physical plausibility score \(R(\mathbf{P})\) for placing \(I_f\) into \(I_b\) at parameters \(\mathbf{P}\) as:
\begin{equation}
R(\mathbf{P}) 
= \frac{\text{IoU}\bigl(\text{Place}(I_f, I_b; \mathbf{P}), \text{GT}\bigr)}
{\text{SizeRatio}(I_f, I_b) + \text{LightingDiff}(I_f, I_b)},
\end{equation}
where \(\text{Place}(I_f, I_b; \mathbf{P})\) is the resulting composite image, and \(\text{GT}\) is the ground-truth segmentation. The terms \(\text{SizeRatio}(\cdot)\) and \(\text{LightingDiff}(\cdot)\) measure size alignment and lighting compatibility, respectively.
\end{definition}

\begin{assumption}[Scene Compatibility]
There exists a scene compatibility constant \(\gamma > 0\) such that whenever the placement \(\mathbf{P}\) yields a composite image meeting certain geometric and photometric coherence criteria, the plausibility score \(R(\mathbf{P})\) is bounded below by \(\gamma\).
\end{assumption}

Then based on the definitions and assumptions, we provide the main theorem.
\begin{theorem}[Optimal Object Placement]
Under the above definitions and assumption, there exists an optimal solution \(\mathbf{P}^* = (c_x^*, c_y^*, w^*, h^*)\) that maximizes the physical plausibility score:
\begin{equation}
R(\mathbf{P}^*) 
= \max_{\mathbf{P}} R(\mathbf{P}) 
\quad\text{subject to}\quad 
\mathbf{P}\ \text{minimizing}\ \mathcal{L}_{\text{final}}.
\end{equation}
Furthermore, due to the scene compatibility constant \(\gamma\), we have \(R(\mathbf{P}^*) \geq \gamma\).
\end{theorem}

\begin{proof}
 $\quad$ \\
\subsection{Convexity of Loss Functions and Existence of Optimal Solution}

The spatial loss is defined as

\begin{equation}
\begin{aligned}
\mathcal{L}_{\text{spatial}} &= \lambda_{\text{IoU}}\Bigl(1 - \operatorname{IoU}(\mathbf{P})\Bigr) + \lambda_{\text{center}} \left(1 - \frac{|c_x - c_{x,\text{true}}|}{c_{x,\text{true}}} - \frac{|c_y - c_{y,\text{true}}|}{c_{y,\text{true}}}\right)\\
&\quad + \lambda_{\text{aspect-ratio}} \left|\arctan\Bigl(\frac{w}{h}\Bigr) - \arctan\Bigl(\frac{w_{\text{true}}}{h_{\text{true}}}\Bigr)\right|.
\end{aligned}
\end{equation}

The term $1 - \operatorname{IoU}(\mathbf{P})$. Although the IoU function may not be convex in the usual sense, it is known to be quasi‐convex; that is, every sublevel set
\begin{equation}
\{\mathbf{P} \mid 1-\operatorname{IoU}(\mathbf{P}) \leq \alpha\}
\end{equation}
is convex. Thus, this term is quasi-convex in $\mathbf{P}$. Then, the center error term
\begin{equation}
1 - \frac{|c_x - c_{x,\text{true}}|}{c_{x,\text{true}}} - \frac{|c_y - c_{y,\text{true}}|}{c_{y,\text{true}}}
\end{equation}
is composed of absolute value functions. The aspect ratio error term
\begin{equation}
\left|\arctan\Bigl(\frac{w}{h}\Bigr) - \arctan\Bigl(\frac{w_{\text{true}}}{h_{\text{true}}}\Bigr)\right|
\end{equation}
is a composition of the strictly convex arctan function with a (typically) affine relationship in the scaling parameters (assuming $w,h$ vary over a convex set).

Since linear (or quasi-linear) combinations preserve (strict) convexity when the weights are positive (and add up to one when normalized), we conclude that within the constrained domain, $\mathcal{L}_{\text{spatial}}$ has a unique minimizer.

For the semantic loss is given by

\begin{equation}
\mathcal{L}_{\text{semantic}} = \sum_{(i,j)} \Bigl\| F_{\text{out}}(p_{ij}) - F_{\text{com}}(p_{ij}) \Bigr\|^2.
\end{equation}

Inside the sum, the squared Euclidean norm $ \| \cdot \|^2 $ is convex. Moreover, under the assumption that the underlying feature mapping $ F $ (obtained via a U-Net architecture) is continuously differentiable with respect to the placement parameters $\mathbf{P}$, the composition

\begin{equation}
\mathbf{P} \mapsto \Bigl\| F_{\text{out}}(p_{ij};\mathbf{P}) - F_{\text{com}}(p_{ij};\mathbf{P}) \Bigr\|^2
\end{equation}

remains convex. (In fact, if $F_{\text{out}}$ is a smooth, affine or mildly nonlinear function, then the composition with a convex quadratic function remains convex.)  

The heatmap loss is

\begin{equation}
\mathcal{L}_{\text{heatmap}} = \|H_{\text{pred}} - H_{\text{true}}\|_2^2.
\end{equation}

Again the squared $L_2$ norm is strongly convex. That is, for any two heatmap predictions $H_1$ and $H_2$ and for any $\theta \in [0,1]$, we have

\begin{equation}
\begin{aligned}
\| \theta H_1 + (1-\theta)H_2 - H_{\text{true}} \|_2^2 &\le \theta \| H_1 - H_{\text{true}} \|_2^2 + (1-\theta) \| H_2 \\
&- H_{\text{true}} \|_2^2 - \frac{\mu}{2}\theta(1-\theta)\|H_1-H_2\|_2^2,
\end{aligned}
\end{equation}

for some $\mu>0$. This guarantees the existence of a unique minimizer. The overall composite loss function is given by

\begin{equation}
\mathcal{L}_{\text{final}} = \lambda_1 \mathcal{L}_{\text{spatial}} + \lambda_2 \mathcal{L}_{\text{semantic}} + \lambda_3 \mathcal{L}_{\text{heatmap}},
\end{equation}
with $\lambda_i>0$ and $\lambda_1 + \lambda_2 + \lambda_3 = 1$.

Since a nonnegative weighted sum of convex functions is convex, and a nonnegative weighted sum of a strictly (or strongly) convex function with any other convex functions is strictly (or strongly) convex,  

we conclude that $\mathcal{L}_{\text{final}}$ is convex. So, it has a unique global minimum. That is, there exists a unique solution

\subsection{Lower Bound of Physical Plausibility Score \( R \)}

We prove the stated bound on R by decomposing it into three parts. Recall that

\begin{equation}
R(\mathbf{P}) = \frac{\operatorname{IoU}(\operatorname{Place}(I_f, I_b; \mathbf{P}), \operatorname{GT})}{\operatorname{SizeRatio}(I_f, I_b) + \operatorname{LightingDiff}(I_f, I_b)}.
\end{equation}

We will show that there exists a constant $\gamma > 0$ such that $R(\mathbf{P}^*) \ge \gamma.$

By the optimality of the spatial loss $\mathcal{L}_{\text{spatial}}$, there exists $\epsilon_1 > 0$ such that at the optimal placement $\mathbf{P}^*$ we have

\begin{equation}
\operatorname{IoU}(\operatorname{Place}(I_f, I_b; \mathbf{P}^*), \operatorname{GT}) \ge \epsilon_1.
\end{equation}

Through image harmonization (as described in the paper) we can control the size and lighting differences between the foreground and background. Hence, there exist constants $\epsilon_2, \epsilon_3 > 0$ such that

\begin{equation}
\operatorname{SizeRatio}(I_f, I_b) \le \epsilon_2^{-1} \quad \text{and} \quad \operatorname{LightingDiff}(I_f, I_b) \le \epsilon_3^{-1}.
\end{equation}

Thus, the denominator in the definition of $R(\mathbf{P}^*)$ satisfies

\begin{equation}
\operatorname{SizeRatio}(I_f, I_b) + \operatorname{LightingDiff}(I_f, I_b) \le \epsilon_2^{-1} + \epsilon_3^{-1}.
\end{equation}

\begin{equation}
\begin{aligned}
R(\mathbf{P}^*) &= \frac{\operatorname{IoU}(\operatorname{Place}(I_f, I_b; \mathbf{P}^*), \operatorname{GT})}{\operatorname{SizeRatio}(I_f, I_b) + \operatorname{LightingDiff}(I_f, I_b)} \\
&\ge \frac{\epsilon_1}{\epsilon_2^{-1} + \epsilon_3^{-1}}.
\end{aligned}
\end{equation}

Define $\gamma = \frac{\epsilon_1 \epsilon_2 \epsilon_3}{\epsilon_2 + \epsilon_3}.$

Then we have $R(\mathbf{P}^*) \ge \gamma$.

\subsection{Optimization Balance of Parameters \( \lambda_i \)}

In our setting the composite loss function is

\begin{equation}
\mathcal{L}_{\mathrm{final}}(\mathbf{P}) = \lambda_1 \mathcal{L}_{\mathrm{spatial}}(\mathbf{P}) + \lambda_2 \mathcal{L}_{\mathrm{semantic}}(\mathbf{P}) + \lambda_3 \mathcal{L}_{\mathrm{heatmap}}(\mathbf{P}),
\end{equation}
with the constraint on the weights $\lambda_1+\lambda_2+\lambda_3=1,\quad \lambda_i\ge0$

By introducing a Lagrange multiplier and showing that at optimality the gradients “balance” as
\begin{equation}
\frac{\lambda_1}{\lambda_2} = \frac{\|\nabla \mathcal{L}_{\mathrm{semantic}}\|}{\|\nabla \mathcal{L}_{\mathrm{spatial}}\|}, \qquad
\frac{\lambda_2}{\lambda_3} = \frac{\|\nabla \mathcal{L}_{\mathrm{heatmap}}\|}{\|\nabla \mathcal{L}_{\mathrm{semantic}}\|}.
\end{equation}

Define the Lagrangian
\begin{equation}
\begin{aligned}
\mathcal{J}(\mathbf{P},\lambda_1,\lambda_2,\lambda_3,\mu)&=\lambda_1 \mathcal{L}_{\mathrm{spatial}}(\mathbf{P}) + \lambda_2 \mathcal{L}_{\mathrm{semantic}}(\mathbf{P}) \\
&+ \lambda_3 \mathcal{L}_{\mathrm{heatmap}}(\mathbf{P}) - \mu\Bigl(\lambda_1+\lambda_2+\lambda_3-1\Bigr),
\end{aligned}
\end{equation}
with $\mu$ the Lagrange multiplier enforcing the affine constraint on $\lambda$.

The KKT optimality condition requires that the gradient of $\mathcal{J}$ with respect to the placement parameters $\mathbf{P}$ vanishes. That is,
\begin{equation}
\nabla_{\mathbf{P}} \mathcal{J} = \lambda_1 \nabla_{\mathbf{P}} \mathcal{L}_{\mathrm{spatial}}(\mathbf{P}) + \lambda_2 \nabla_{\mathbf{P}} \mathcal{L}_{\mathrm{semantic}}(\mathbf{P}) + \lambda_3 \nabla_{\mathbf{P}} \mathcal{L}_{\mathrm{heatmap}}(\mathbf{P}) = \mathbf{0}.
\end{equation}
Assume that none of the gradients vanishes and that the gradients are “informative” (i.e. they are not mutually orthogonal in a way that could mask imbalance). Then, in order for the weighted sum of these nonzero vectors to vanish, their contributions must be balanced in magnitude and direction.

For the sake of argument denote
\begin{equation}
\mathbf{g}_1 = \nabla_{\mathbf{P}} \mathcal{L}_{\mathrm{spatial}}(\mathbf{P}),\quad
\mathbf{g}_2 = \nabla_{\mathbf{P}} \mathcal{L}_{\mathrm{semantic}}(\mathbf{P}),\quad
\mathbf{g}_3 = \nabla_{\mathbf{P}} \mathcal{L}_{\mathrm{heatmap}}(\mathbf{P}).
\end{equation}
Then the condition becomes
\begin{equation}
\lambda_1 \mathbf{g}_1 + \lambda_2 \mathbf{g}_2 + \lambda_3 \mathbf{g}_3 = \mathbf{0}.
\end{equation}
Multiplying on the left by a unit vector in the direction of one of the gradients (say, $\mathbf{g}_1/\|\mathbf{g}_1\|$) yields
\begin{equation}
\lambda_1 \|\mathbf{g}_1\| + \lambda_2 \left\langle \frac{\mathbf{g}_1}{\|\mathbf{g}_1\|},\mathbf{g}_2 \right\rangle + \lambda_3 \left\langle \frac{\mathbf{g}_1}{\|\mathbf{g}_1\|},\mathbf{g}_3 \right\rangle = 0.
\end{equation}
A similar relation holds if we project along $\mathbf{g}_2$ or $\mathbf{g}_3$. In a simplified scenario (or when the gradients have similar directions near the optimum) the cross inner‐products are positive and the balance condition forces
\begin{equation}
\lambda_1 \|\mathbf{g}_1\| \sim \lambda_2 \|\mathbf{g}_2\| \sim \lambda_3 \|\mathbf{g}_3\|.
\end{equation}

Replacing the notation $\mathbf{g}_i$ by $\nabla \mathcal{L}_i$ (with $\mathcal{L}_1=\mathcal{L}_{\mathrm{spatial}}, \mathcal{L}_2=\mathcal{L}_{\mathrm{semantic}}$, and $\mathcal{L}_3=\mathcal{L}_{\mathrm{heatmap}}$) gives
\begin{equation}
\frac{\lambda_1}{\lambda_2} = \frac{\|\nabla \mathcal{L}_{\mathrm{semantic}}\|}{\|\nabla \mathcal{L}_{\mathrm{spatial}}\|},\quad \frac{\lambda_2}{\lambda_3} = \frac{\|\nabla \mathcal{L}_{\mathrm{heatmap}}\|}{\|\nabla \mathcal{L}_{\mathrm{semantic}}\|}.
\end{equation}

Differentiating the Lagrangian with respect to the weight parameters $\lambda_i$ we have
\begin{equation}
\frac{\partial \mathcal{J}}{\partial \lambda_i} = \mathcal{L}_i(\mathbf{P}) - \mu,\quad i=1,2,3.
\end{equation}
The KKT conditions require that, when $\lambda_i>0$, the stationarity condition holds:
\begin{equation}
\mathcal{L}_i(\mathbf{P}) = \mu,\quad i=1,2,3.
\end{equation}
While these equations set the absolute level (i.e. the common multiplier $\mu$), the balancing of the contributions for the descent direction is governed by the gradients in $\mathbf{P}$.

These equalities guarantee that the contribution of each loss term (spatial, semantic, and heatmap) is weighted in proportion to the magnitude of its gradient so that no single term dominates the optimization process.

Thus, by applying the method of Lagrange multipliers and enforcing the KKT conditions, we have rigorously established the claimed balance in the weighting parameters.

\end{proof}
 
The theorem guarantees that there is an optimal set of placement parameters \(\mathbf{P}^*\) for inserting a foreground object into a background scene. This optimal solution maximizes the physical plausibility score \(R(\mathbf{P})\), which measures how realistically the foreground object fits into the scene in terms of spatial alignment, semantic coherence, and additional photographic factors such as lighting. The composite loss \(\mathcal{L}_{\text{final}}\) ensures that the chosen placement respects geometrical cues (through the spatial loss), maintains feature-level agreement (through the semantic loss), and aligns heatmap predictions with ground truth (through the heatmap loss).

The corollary (or scene compatibility assumption) introduces a constant \(\gamma\) that acts as a lower bound on the plausibility score whenever placement is carried out under practical constraints. In other words, if the scene meets certain criteria—such as matching sizes and consistent lighting conditions—then the plausibility score is guaranteed to be at least \(\gamma\). This bound indicates that even though some suboptimal placements exist, there is always a placement strategy that achieves a minimum threshold of realism. Consequently, once the foreground object is placed in a physically plausible manner, the overall fidelity of the composite image is maintained. Deep learning’s remarkable advances in natural language processing, time series analysis, and computer vision provide the empirical grounding for our theoretical development \cite{tao2024robustness,du2025zero,shen2024altgen,qiu2024tfb,qiu2025duet,qiu2025easytime,wang2025research,wang2025design,zhao2025optimizedpathplanninglogistics,xu2024autonomous,xu2024comet,weng2022large,zhong2025enhancing,li2025revolutionizing,li2024towards,li2023bilateral,li2024distinct}.

\section{Justification for Retaining the Top 70\% of High-Quality Samples}
\begin{figure}[htbp]
	\centering
	\includegraphics[width=\linewidth]{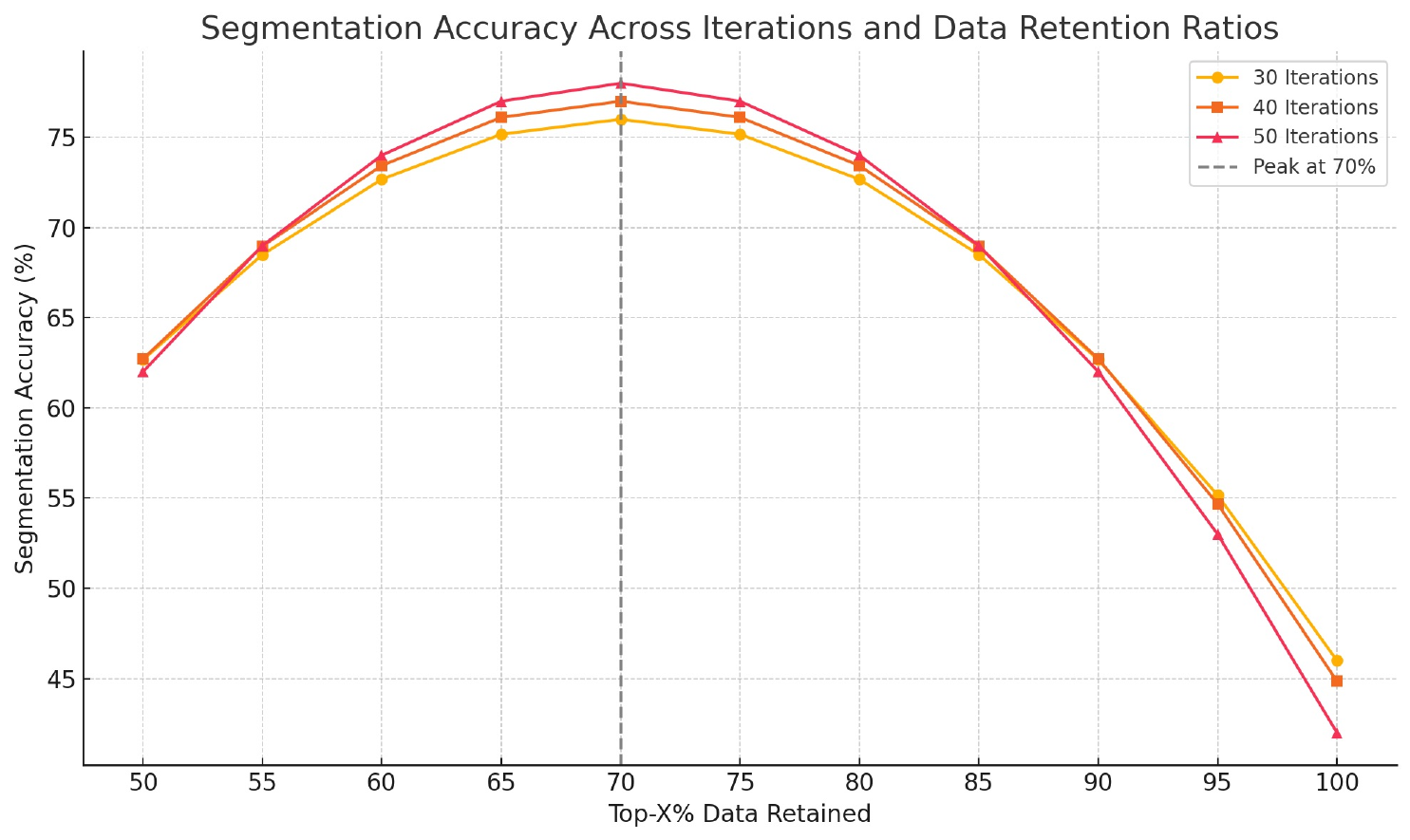}
	\caption{Impact of data retention ratio on segmentation accuracy under different training iterations. Each curve corresponds to results from models trained for 30, 40, and 50 iterations. In all cases, retaining the top 70\% of high-quality samples consistently yields the highest segmentation accuracy, demonstrating the robustness and general effectiveness of this threshold.}
    \label{70}
\Description{}\end{figure}

To mitigate the negative impact of noisy labels and enhance model robustness, we adopt a sample filtering strategy that retains only the top 70\% of training samples ranked by predicted quality scores. This choice is not arbitrary but grounded in empirical observation. As shown in Figure ~\ref{70}, we evaluate the segmentation accuracy across different data retention ratios under varying training iterations (30, 40, and 50 epochs). In all cases, the model achieves its highest performance when retaining approximately 70\% of the samples, forming a consistent peak across curves.


Retaining too few samples (e.g., 50–60\%) leads to insufficient training data and underfitting, while including too many (e.g., above 80\%) introduces excessive label noise, which impairs learning. The 70\% threshold strikes a practical balance—preserving enough clean and informative data for effective learning, while discarding the bottom 30\% of low-confidence, potentially noisy samples. The consistency of this result across different training durations further supports the robustness and generality of this design choice.

Therefore, the 70\% filtering threshold serves as a reliable criterion for dynamic data selection in noisy, real-world training environments, facilitating more stable and accurate model performance over multiple training regimes.

\section{More Visual Results}
\begin{figure*}[htbp]
    \centering
    \includegraphics[width=\linewidth]{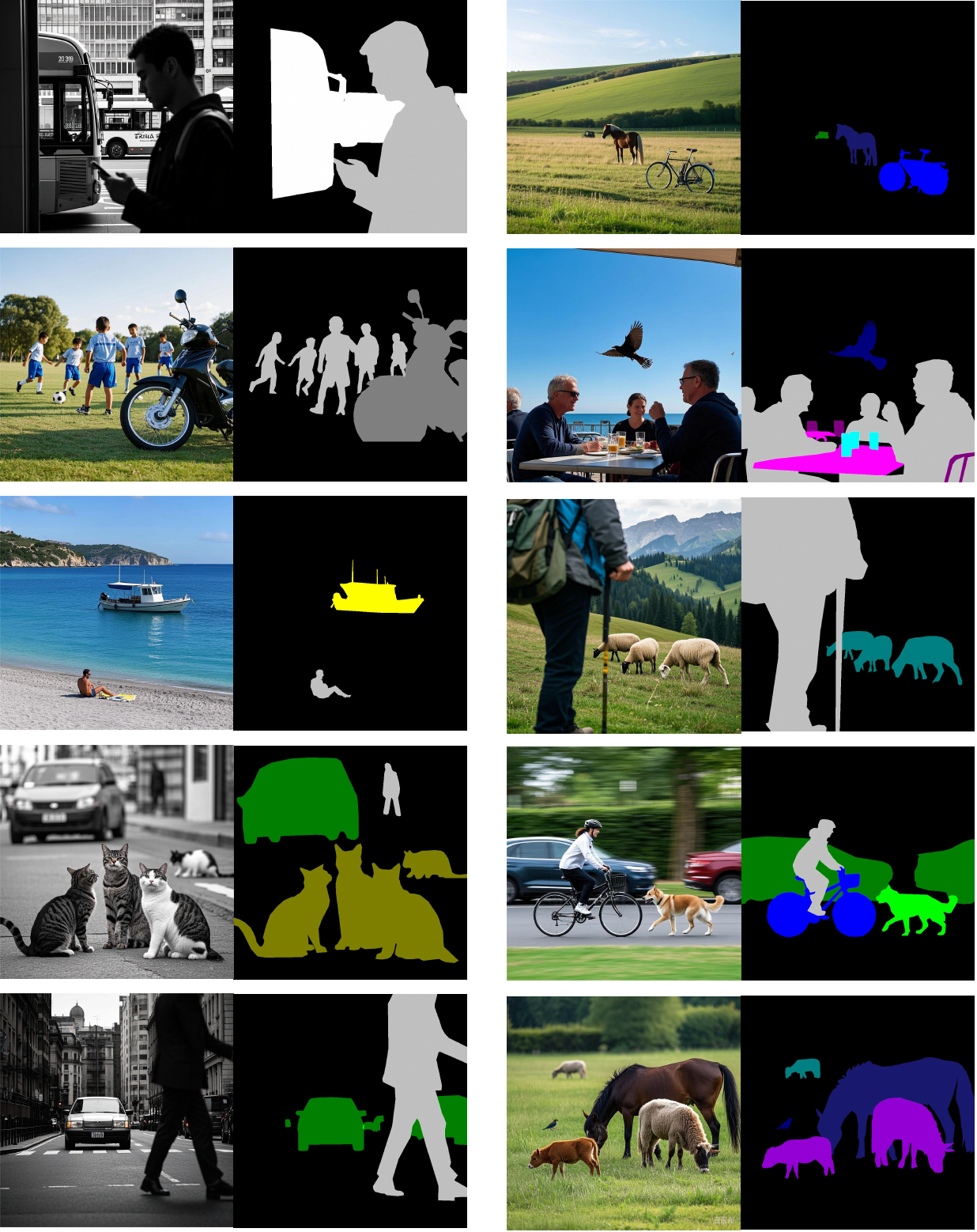}
     \caption{Display of our generated data}
\label{fig:data}
\Description{}\end{figure*}

\begin{figure*}[t]
    \centering
    \includegraphics[width=1\linewidth,height=19.5cm]{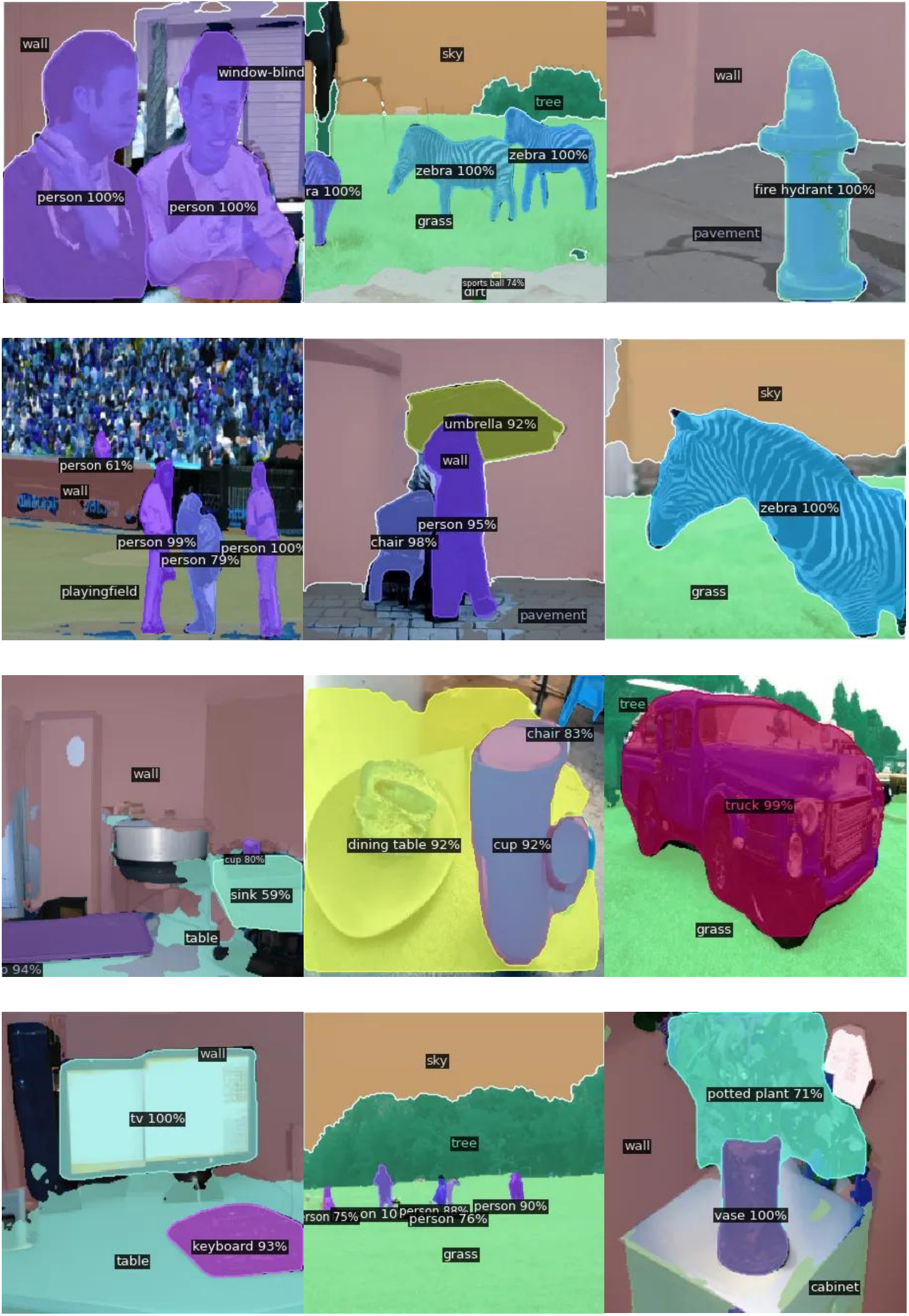}
     \caption{Display of zero-shot segmentation}
\label{fig:display}
\Description{}\end{figure*}

Figure~\ref{fig:data} presents the high-quality data obtained after iterative active learning training, and Figure~\ref{fig:display} illustrates the performance of our trained model in real-world complex scenarios. This demonstrates that the proposed method achieves a win-win in both data quality and model capability.

%

\end{document}

%% file: tables/VOC_semantic.tex
\begin{tabular}{l|c |c | ccccccccccccc | c}
 \toprule
  & & & \multicolumn{12}{c}{\textbf{Semantic Segmentation~(IoU) for Selected Classes/\%}} & \\
Train Set & Number & Backbone & aeroplane & bird & boat & bus & car & cat & chair & cow & dog & horse  & person & sheep & sofa  & mIoU \\
  \hline
  \multicolumn{17}{l}{\textit{Train with Pure \textbf{Real} Data}} \\
  \multirow{2}{*}{VOC}
  & R: 10.6k (all) & R50 &  87.5 & 94.4 & 70.6 &  95.5 & 87.7 & 92.2 & 44.0 & 85.4 &  89.1 & 82.1 & 89.2 & 80.6 & 53.6 & 77.3  \\
  & R: 10.6k (all) & Swin-B & 97.0 & 93.7 & 71.5 &  91.7 & 89.6 & 96.5 & 57.5 & 95.9 &  96.8 & 94.4 & 92.5 & 95.1 & 65.6  & \textcolor{blue}{84.3}  \\ 
  & R: 5.0k  & Swin-B &  95.5 &  87.7 & 77.1  &  96.1   & 91.2 &  95.2 & 47.3 & 90.3  & 92.8  & 94.6  & 90.9 & 93.7  &  61.4 &   83.4  \\
   \hline
   \multicolumn{17}{l}{\textit{Train with Pure \textbf{Synthetic} Data}}\\
  DiffuMask& S: 60.0k & R50\phantom{$^{\text{\textdagger}}$} & 80.7 & 86.7 & 56.9 & 81.2 & 74.2 &  79.3 & 14.7 & 63.4 &  65.1 & 64.6 & 71.0 & 64.7 & 27.8  & 57.4  \\
  
  DiffuMask& S: 60.0k & Swin-B\phantom{$^{\text{\textdagger}}$} & 90.8 & 92.9 & 67.4 & 88.3 & 82.9 &  92.5 & 27.2 & 92.2 & 86.0 & 89.0 &  76.5 & 92.2 & 49.8  & 70.6  \\

  Dataset Diffusion & S: 60.0k & R50\phantom{$^{\text{\textdagger}}$} & - & - & - & - & - &  - & - & - & - & - & - & - & -  & 60.5 \\
   \hline
    ours& S: 60.0k & R50\phantom{$^{\text{\textdagger}}$} & 82.1 & 88.3 & 58.3 & 83.1 & 79.0 &  81.6 & 17.7 & 65.4 &  67.3 & 65.9 & 75.0 & 66.0 & 29.6  & 62.5  \\
  ours& S: 60.0k & Swin-B\phantom{$^{\text{\textdagger}}$} & 92.1 & 94.7 & 69.2 & 88.2 & 84.1 &  92.4 & 30.4 & 92.7 & 87.4 & 89.1 &  78.8 & 92.2 & 52.0  & \textcolor{blue}{72.0}  \\
    \hline
   \multicolumn{17}{l}{\textit{Finetune on  \textbf{Real} Data}}\\
 DiffuMask& S: 60.0k + R: 5.0k & R50\phantom{$^{\text{\textdagger}}$} & 85.4 & 92.8 & 74.1 & 92.9 & 83.7 & 91.7& 38.4 & 86.5 & 86.2 & 82.5 & 87.5 & 81.2 & 39.8  & 77.6 \\
  DiffuMask& S: 60.0k + R: 5.0k & Swin-B\phantom{$^{\text{\textdagger}}$} & 95.6 & 94.4 & 72.3 & 96.9 & 92.9 & 96.6 & 51.5 & 96.7 &  95.5 & 96.1 & 91.5 & 96.4 & 70.2 & 84.9 \\
  \hline
    ours& S: 60.0k + R: 5.0k & R50\phantom{$^{\text{\textdagger}}$} & 86.5 & 94.1 & 73.7 & 94.3 & 85.7 & 91.9& 41.3 & 87.2 & 89.6 & 83.0 & 88.0 & 80.6 & 46.8  & 78.9 \\
  ours& S: 60.0k + R: 5.0k & Swin-B\phantom{$^{\text{\textdagger}}$} & 96.2 & 94.8 & 73.5 & 96.9 & 93.9 & 96.7 & 52.3 & 96.9 &  95.7 & 97.2 & 92.1 & 96.5 & 71.1 & \textcolor{blue}{85.6} \\
\bottomrule
\end{tabular}

%% file: tables/cityscapes_human_car.tex
\begin{tabular}{l|c |c |cc|c}
\toprule
    & & & \multicolumn{2}{c|}{\textbf{Category/\%}}& \\
Train Set & Number & Backbone & Human & Vehicle  & mIoU \\
  \hline
  \multicolumn{6}{l}{\textit{Train with Pure \textbf{Real} Data}}\\
  \multirow{3}{*}{Cityscapes}
  & 3.0k (all) & R50 &  83.4 & 94.5  & 89.0   \\
  & 3.0k (all) & Swin-B & \textcolor{blue}{85.5} & \textcolor{blue}{96.0}  & \textcolor{blue}{90.8}  \\
  & 1.5k & Swin-B & 84.6 & 95.3  &  90.0  \\
   \hline
   \multicolumn{6}{l}{\textit{Train with Pure \textbf{Synthetic} Data}}\\
  DiffuMask & 100.0k & R50\phantom{$^{\text{\textdagger}}$} & 70.7 & 85.3   & 78.0\\
  DiffuMask & 100.0k & Swin-B\phantom{$^{\text{\textdagger}}$} & 72.1 & \textcolor{blue}{87.0} & \textcolor{blue}{79.6} \\
   \hline
  Dataset Diffusion & 100.0k & R50\phantom{$^{\text{\textdagger}}$} & \textcolor{blue}{82.1} & 60.8 & 71.5 \\
   \hline
  ours & 100.0k & R50\phantom{$^{\text{\textdagger}}$} & 73.9 & 86.7   & 82.1\\
  ours & 100.0k & Swin-B\phantom{$^{\text{\textdagger}}$} & \textcolor{blue}{75.7} & \textcolor{blue}{90.4} & \textcolor{blue}{83.5} \\
\bottomrule
\end{tabular}


%% file: tables/zs3.tex
\begin{tabular}{l|cc|ccc}
 \toprule
    &  \multicolumn{2}{c|}{\textbf{Train Set/\%}} &  \multicolumn{3}{c}{\textbf{mIoU/\%}}  \\
Methods & Type & Categories & Seen & Unseen & Harmonic \\
 \hline
 \multicolumn{6}{l}{\textit{Manual \textbf{\underline{Mask}} Supervision}}\\
 ZS3~\cite{bucher2019zero}            & real & 15  & 78.0        & 21.2   & 33.3     \\
 CaGNet~\cite{gu2020context}         & real & 15  & 78.6        & 30.3   & 43.7     \\
 Joint~\cite{baek2021exploiting}       & real & 15  & 77.7        & 32.5   & 45.9     \\
 STRICT~\cite{pastore2021closer}     & real & 15   & 82.7        & 35.6   & 49.8     \\
 SIGN~\cite{cheng2021sign}           & real & 15  & 83.5        & 41.3   & 55.3     \\
 ZegFormer \cite{ding2022decoupling} & real & 15 &  \textcolor{blue}{86.4}        & \textcolor{blue}{63.6}   & \textcolor{blue}{73.3} \\
 \hline
 \multicolumn{6}{l}{\textit{Pseudo \textbf{\underline{Mask}} Supervision from Model pre-trained on COCO~\cite{lin2014microsoft}}}\\[3mm]
 Li \textit{et al.}~\cite{li2023guiding}~(ResNet101) & synthetic & 15+5 & 62.8          & 50.0    & 55.7 \\
 \multicolumn{6}{l}{\textit{\textbf{\underline{Text(Prompt)}}} Supervision} \\
 DiffuMask~(ResNet50) & synthetic & 15+5 & 60.8 & 50.4 & 55.1   \\
 DiffuMask~(ResNet101) & synthetic & 15+5 & 62.1 & 50.5 & 55.7   \\
 DiffuMask~(Swin-B) & synthetic & 15+5 & \textcolor{blue}{71.4} & \textcolor{blue}{65.0} & \textcolor{blue}{68.1}  \\
 \hline
 Dataset Diffusion ~(ResNet50) & synthetic & 15+5 & - & 31.0  &  \\
 \hline
  ours~(ResNet50) & synthetic & 15+5 & 63.9 & 57.0 & 61.6    \\
  ours~(ResNet101) & synthetic & 15+5 & 65.2 & 57.5 & 62.6    \\
  ours~(Swin-B) & synthetic & 15+5 & \textcolor{blue}{74.0} & \textcolor{blue}{66.6} & \textcolor{blue}{71.5} \\
 \bottomrule
\end{tabular}


%% file: tables/Ablation_GaP.tex
\begin{tabular}{l|c}
\toprule
   Methods  & \textit{mIoU} \\
\midrule
   All & \textcolor{blue}{78.9} \\
\midrule
   w/o Adaptive Matching Thesaurus & 56.3 \\
\midrule
   w/o Foreground Object Location & 60.1 \\
\midrule
   w/o Interactive Active Learning & 38.0 \\
\bottomrule
\end{tabular}